\documentclass[10pt,twocolumn,letterpaper]{article}

\usepackage{verbatim}
\usepackage{booktabs}
\usepackage{multirow}
\usepackage{colortbl}
\usepackage{xcolor}
\usepackage{algorithm}
\usepackage{gensymb}
\usepackage[noend]{algpseudocode}
\usepackage{graphicx}
\usepackage[pagenumbers]{cvpr} 

\definecolor{cvprblue}{rgb}{0.21,0.49,0.74}
\usepackage[pagebackref,breaklinks,colorlinks,allcolors=cvprblue]{hyperref}

\def\confName{CVPR}
\def\confYear{2026}

\title{N4MC: Neural 4D Mesh Compression}

\author{Guodong Chen\\
Northeastern University\\
{\tt\small chen.guod@northeastern.edu}
\and
Huanshuo Dong\\
University of Science and Technology of China\\
{\tt\small bingo000@mail.ustc.edu.cn}
\and
Mallesham Dasari\\
Northeastern University\\
{\tt\small m.dasari@northeastern.edu}
}

\begin{document}

\twocolumn[{%
\renewcommand\twocolumn[1][]{#1}%
\maketitle
\centering
\vspace{-0.3in}
\includegraphics[width=\linewidth]{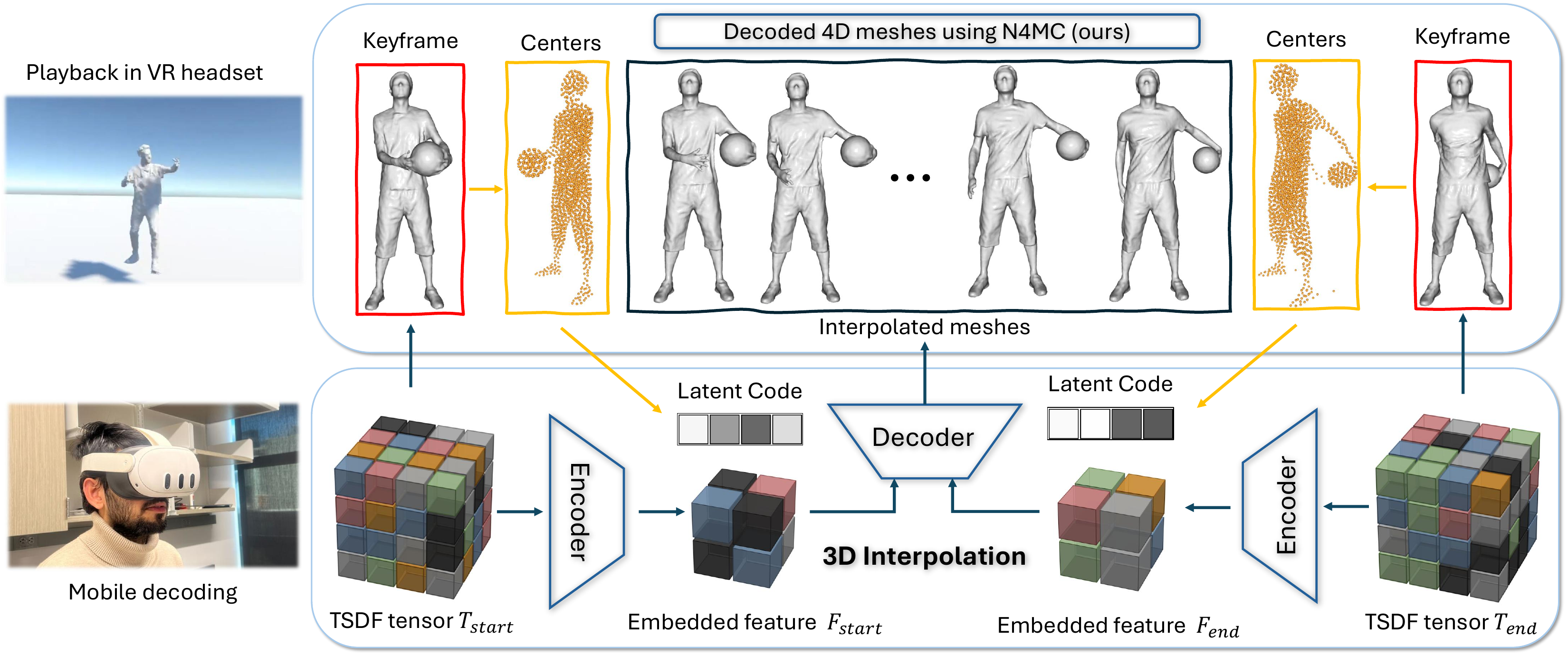}
\captionof{figure}{N4MC is the first neural 4D mesh compression framework. It first converts 4D meshes into a modified version of Truncated Signed Distance Function (TSDF) tensors and compresses them through an auto-encoder-decoder pair. A volume tracking-based transformer model exploits temporal redundancy by learning 3D interpolation across frames. N4MC produces lightweight, sequence-specific models optimized for each mesh sequence, enabling real-time decoding on desktops. N4MC also provides a Unity plugin for mobile devices (e.g., Meta Quest 3) decoding and playback.
\vspace{1em}} 
\label{fig:teaser}
}]

\begin{abstract} 
We present N4MC, the first 4D neural compression framework to efficiently compress time-varying mesh sequences by exploiting their temporal redundancy. Unlike prior neural mesh compression methods that treat each mesh frame independently, N4MC takes inspiration from inter-frame compression in 2D video codecs, and learns motion compensation in long mesh sequences. Specifically, N4MC converts consecutive irregular mesh frames into regular 4D tensors to provide a uniform and compact representation. These tensors are then condensed using an auto-decoder, which captures both spatial and temporal correlations for redundancy removal. To enhance temporal coherence, we introduce a transformer-based interpolation model that predicts intermediate mesh frames conditioned on latent embeddings derived from tracked volume centers, eliminating motion ambiguities. Extensive evaluations show that N4MC outperforms state-of-the-art in rate-distortion performance, while enabling real-time decoding of 4D mesh sequences. 
The implementation of our method is available at:
\url{https://github.com/frozzzen3/N4MC}.

\end{abstract}    
\vspace{-0.1in}
\section{Introduction}
\label{sec:intro}

Triangle meshes are among the most established and widely adopted representations of 3D geometry in AR/VR, robotics, and digital twin systems. As 3D capture and reconstruction technologies mature, the scale and complexity of these meshes have grown rapidly, with high-resolution models easily reaching millions of vertices per frame and hundreds of frames in time-varying (dynamic) sequences. This surge in data volume poses fundamental challenges for storage and transmission without compromising quality.

We propose N4MC, a neural framework that compresses long time-varying 3D mesh sequences by learning to interpolate geometry over time rather than explicitly encoding every frame. N4MC achieves high compression ratios while maintaining high visual fidelity by exploiting temporal redundancy across consecutive mesh frames. 


Traditional mesh compression methods like TFAN~\cite{mamou2009tfan} and EdgeBreaker~\cite{rossignac1999edgebreaker} achieve excellent compression on a per-frame basis. While these methods are popular even today, they do not exploit the temporal redundancy between consecutive frames, resulting in high data rates for long mesh sequences. Recent methods~\cite{vasa2007coddyac, amjoun2009single, luo2020spatio, apple2022VSMC} and MPEG standards such as IC~\cite{bourges2002introduction}, MESHGRID~\cite{salomie2004meshgrid}, and FAMC~\cite{mamou2006skinning} explored exploiting spatial and temporal correlations between frames, but focus mainly on meshes with consistent topology. Even proposals like Apple’s VSMC~\cite{apple2022VSMC} adopted for MPEG's V-DMC standard~\cite{V-DMC}, depends on temporally consistent re-meshing that is often impractical. Deformation methods~\cite{jin2024embedded, hoang2023embedded, sorkine2007rigid, sumner2007embedded} morph a reference mesh into target frames, but struggle with non-rigid motions and self-contact issues. Latest work TVMC~\cite{chen2025tvmc} uses volume center affinity to exploit temporal redundancy but struggles with non-rigid motions, leading to deformation breakdowns. 



Neural methods have shown strong potential for 3D geometry compression. For example, neural progressive meshes~\cite{chen2023neural} combine multiresolution and level of detail (LoD) concepts for scalable reconstruction. DMGC~\cite{zhao2023dmgc} predicts mesh vertices and connectivity with graph-based models often struggle with large vertex counts. Similarly, methods such as NeCGS~\cite{ren2024necgs}, Squeeze3D~\cite{dagli2025squeeze3d}, and others~\cite{neuralgeometry2024} explore implicit and generative representations to achieve compact encodings of static meshes. While these approaches yield impressive compression for individual models, they do not exploit the temporal redundancy inherent in time-varying mesh sequences.


N4MC is designed to compress long time-varying mesh sequences, addressing the limitations of per-frame and topology-constrained methods. N4MC stores 4D geometry in a compact latent space through a series of learning stages. First, explicit meshes are converted into implicit 4D tensors to provide a uniform volumetric representation of the geometry. Second, these tensors are encoded into compact latent embeddings using a trained autoencoder that captures spatial and temporal correlations jointly. Third, we introduce volume tracking across the mesh sequence to estimate consistent and stable volume centers that serve as motion priors to resolve temporal ambiguities. Finally, a lightweight transformer predicts intermediate-frame embeddings conditioned on key-frame features and volume priors, enabling temporal super-resolution in the latent domain.

We evaluate N4MC on a diverse suite of 4D mesh sequences, including real and synthetic data and both single-object and multi-object scenes. Extensive qualitative and quantitative experiments show that N4MC outperforms state-of-the-art compression performance while maintaining high visual quality. Notably, N4MC can efficiently handle mesh sequences of over 100 frames, while most existing time-varying mesh compression methods are limited to only 5 to 10 frames per group in inter-frame prediction.

Our main contributions are:
\begin{enumerate}
    \item We propose the first neural framework explicitly designed for 4D mesh compression, introducing a tensor-based latent interpolation paradigm that leverages spatiotemporal correlations across mesh sequences to achieve extreme mesh compression efficiency.
    \item We introduce explicit volume center priors to guide interpolation, which accelerates model convergence and helps the transformer to remain lightweight.
    \item We provide an N4MC Unity plugin implementation that enables decoding and playback on VR headsets such as Meta Quest 3 and Android smartphones.
\end{enumerate}

\begin{figure*}[t]
    \centering
    \includegraphics[width=1\linewidth]{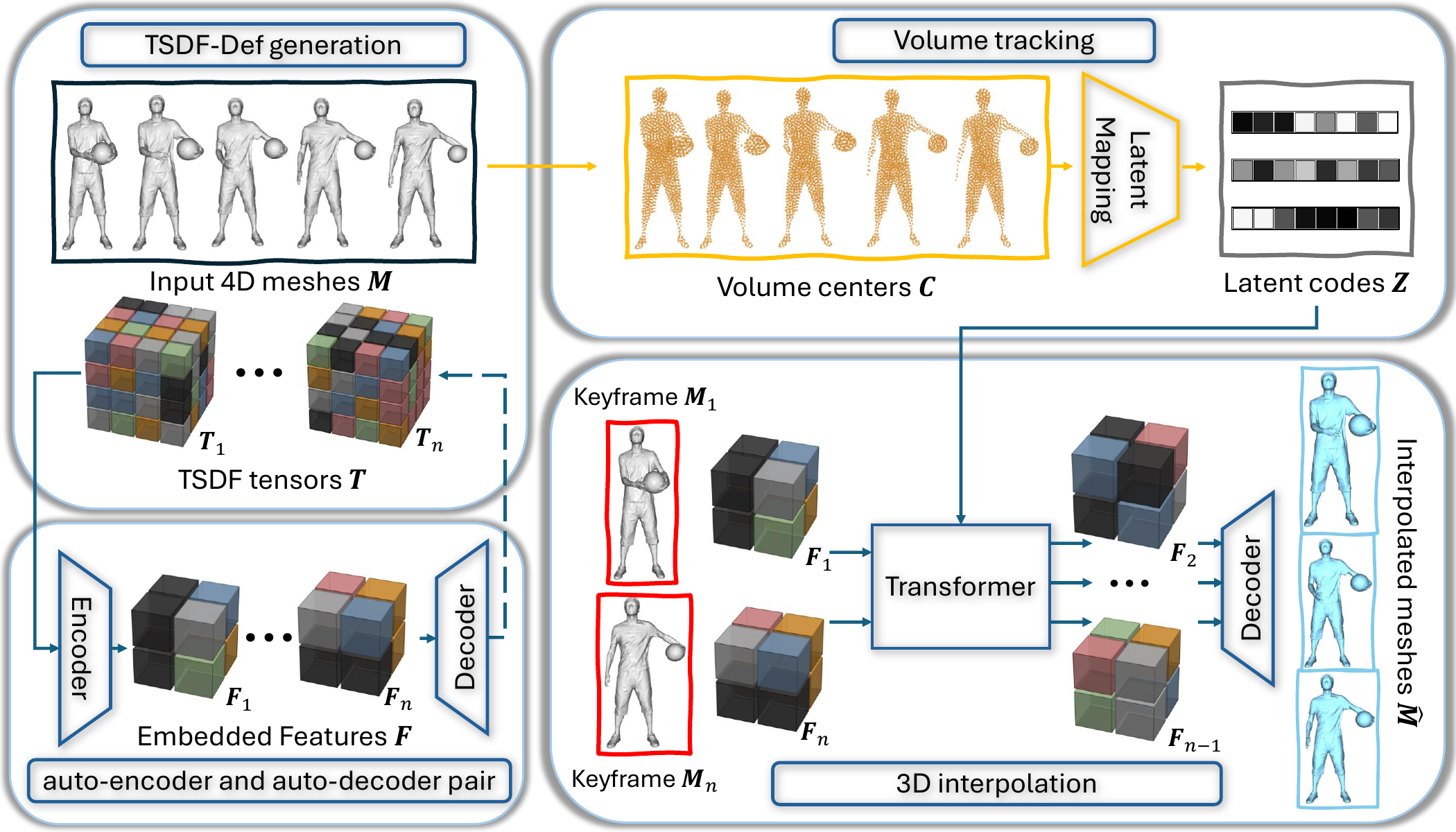} 
    \caption{N4MC overview. N4MC enables neural 4D mesh compression with 4 modules: (1) A TSDF-Def generation module (top left); (2) A trained auto-encoder and auto-decoder pair to condense TSDF-Def tensors (bottom left); (3) A volume tracking module and a latent code mapping network to generate interpolation priors (top right); (4) A light-weight transformer model for 3D interpolation (bottom right).} \vspace{-0.1in}
    \label{fig:workflow}
\end{figure*}

\section{Related Work}
\label{sec:relatedwork}

\paragraph{Static mesh compression.} Static mesh compression utilizes intra-frame redundancy and performs quantization to compress each mesh independently~\cite{deering1995geometry, taubin1998geometric, touma1998triangle, rossignac1999edgebreaker, mamou2009tfan, Draco2024}. These methods usually reorder the mesh structure to enable efficient lossless compression. However, much of this work (e.g., Draco~\cite{Draco2024}, TFAN~\cite{mamou2009tfan}), operate purely on individual frames and lacks inter-frame prediction, making it less efficient for compressing time-varying scene meshes. 

Several lossy compression methods based on implicit mesh representations have been proposed as promising alternatives~\cite{realtime2018, ren2024necgs}. These methods convert explicit 3D meshes into implicit representations such as Signed Distance Fields (SDFs) and Truncated Signed Distance Fields (TSDFs), and then compress the resulting TSDF tensors. These methods too treat TSDF tensors independently and do not exploit the temporal redundancy in consecutive time-varying mesh frames.

\vspace{-0.15in}
\paragraph{Dynamic \& time-varying mesh compression.} Dynamic mesh compression usually encodes a sequence of meshes by storing a base mesh and a displacement field, which is feasible because of their constant topology~\cite{isoCfp, apple2022VSMC}. However, dynamic mesh compression depends on topology consistency across the input mesh sequence. To relax this constraint for time-varying meshes, Apple VSMC~\cite{apple2022VSMC} introduces a time-consistent re-meshing method. Nevertheless, its instability often leads to inaccurate inter-frame correspondences, making the representation of inter-frame differences between time-varying meshes still challenging.

In contrast, time-varying mesh compression deals with mesh sequences where both the geometry and the topology evolve over time (typically captured from the real-world depth sensors)~\cite{han2007time, han2008geometry, yamasaki2010patch, doumanoglou2014toward, hou2014novel, hou2014highly, graziosi2021video, marvie2022compression, hoang2023embedded, MPEGtmc2, jin2024embedded, chen2025tvmc}, making it a more complex problem to solve. A common strategy in the prior work is to transform time-varying meshes into dynamic meshes through deformation or registration. However, such methods are prone to visible distortion and can fail under non-rigid motion, limiting their applicability in real-world meshes extracted from depth sensor data.

\vspace{-0.15in}
\paragraph{Neural mesh compression.} Neural methods achieved remarkable progress in 2D video compression~\cite{wu2018video, chen2021nerv, chen2023hnerv, mentzer2019practical}. Building on this, recent neural methods explored 3D data compression, both in explicit~\cite{lin2022dmvc, zhao2023dmgc} and implicit representations such as SDFs, or volumetric occupancy grids~\cite{ren2024necgs, mescheder2019occupancy, park2019deepsdf, takikawa2021neural, wang2021neus, molnar2024variational, zhao2023michelangelo, peng2020convolutional}, including emerging radiance fields~\cite{takikawa2022variable, chen2022tensorf, chen2023dictionary}. Sequeeze3D~\cite{dagli2025squeeze3d} uses pretrained encoders and generative models to store 3D data in a compact latent code. Neural progressive meshes~\cite{chen2023neural} learns LoD while DMGC~\cite{zhao2023dmgc} predicts vertices and connectivity with graph-based models. However, all of these methods mainly target individual mesh frame compression, limiting their efficiency in long mesh sequences.

NeCGS~\cite{ren2024necgs} is the most closely related to our work, which converts 3D meshes into an enhanced variation of TSDF called TSDF-Def tensors and trains an auto-decoder to compress them into compact embedded features. However, NeCGS is not designed to exploit temporal redundancy in long mesh sequences, and it only captures limited random redundancy among embedded features due to the restricted capacity of its auto-decoder. Compared with these methods, our method extends interpolation to 3D meshes, enabling efficient compression of long mesh sequences (e.g., over 100 frames) by fully exploiting temporal redundancy between adjacent frames.

\section{Proposed Method: N4MC}


\subsection{Preliminary}

N4MC builds on recent shape extracting
methods~\cite{shen2023flexible, shen2021deep} and NeCGS~\cite{ren2024necgs}, which first converts mesh frames into 4D TSDF-Def tensors and then trains an auto-decoder to compress TSDF-Def tensors into compact embedded features.

\vspace{-0.1in}
\paragraph{TSDF-Def.} Inspired from NeCGS~\cite{ren2024necgs}, N4MC adopts an enhanced signed distance representation called TSDF-Def, which embeds both geometric structure and local deformation within a unified volumetric tensor. 
Specifically, a 3D mesh is represented as a 4D tensor $\mathbf{T} \in \mathbb{R}^{k\times k\times k \times 4}$, where k denotes the voxel grid resolution. The last dimension consists of one TSDF value, representing the signed distance from this grid point to the mesh surface, and a corresponding deformation vector $(\Delta x, \Delta y,\Delta z)$, representing the deformation of this grid point in 3D space. 

\subsection{N4MC Overview}

\cref{fig:workflow} shows an overview of N4MC which consists of four main components: (1) a TSDF-Def generation module, which converts an $N$-frame 3D mesh sequence $\mathcal{M}=\{\mathbf{M}_i\}_{i=1}^{N}$ into a series of TSDF-Def tensors $\mathcal{T}=\{\mathbf{T}_i\}_{i=1}^{N}$. (2) a trained auto-encoder $E_\theta(\cdot)$ and auto-decoder $D_\psi(\cdot)$ pair, which compresses TSDF tensors into compact embedded features. (3) a volume tracking module, which tracks volume centers $\mathcal{C}=\{\mathbf{C}_i\}_{i=1}^{N}$ across the sequence and uses a latent code mapping network $f_\omega(\cdot)$ to generate priors that guide the transformer model and reduce ambiguity. (4) a light-weight transformer model, which predicts intermediate frame embedded features using key frame features and latent codes derived from the tracked volume centers. During decoding, the highly compressed latent vectors, together with the lightweight auto-decoder and transformer model, enable efficient streaming and reconstruction of high-quality mesh sequences.

\subsection{Latent Codes Generation}

Generating stable and expressive latent codes for time-varying 3D meshes is fundamentally challenging. Mesh sequences often exhibit irregular topology, varying surface visibility, non-uniform motion, and fine-grained local deformations, all of which make it difficult to derive a compact representation that remains consistent across time.

\vspace{-0.1in}
\paragraph{Auto-encoder.} To obtain latent codes that are both compact and temporally coherent, we introduce a 3D convolutional auto-encoder designed specifically for TSDF-Def tensors. The encoder $E_\theta(\cdot)$ employs a ConvNeXt3D~\cite{liu2022convnet} backbone to extract spatially structured volumetric features from the input TSDF-Def tensors $\mathbf{T}$:
\begin{equation}
    \mathbf{F} = E_\theta(\mathbf{T}),
\end{equation}
where $\mathbf{F}\in \mathbb{R}^{k^{\prime}\times k^{\prime}\times k^{\prime} \times d}$ is the reconstructed TSDF-Def tensors. N4MC is designed for time-varying mesh sequences, where all the meshes have similar motion patterns despite variations in topology and connectivity. This auto-encoder enforces structural consistency across frames so that embedded features $\mathcal{F}=\{\mathbf{F}_i\}_{i=1}^N$ of temporally adjacent meshes remain semantically aligned and predictable.

\vspace{-0.1in}
\paragraph{Auto-decoder.} \label{auto-decoder} N4MC employs a quantization-aware auto-decoder that compresses 4D TSDF-Def tensors into low-dimensional latent embeddings. For a TSDF-Def tensor $\mathbf{T}$, the corresponding input for the auto-decoder is the embedded features $\mathbf{F}\in \mathbb{R}^{k^{\prime}\times k^{\prime}\times k^{\prime} \times d}$ from auto-decoder, where $k^{\prime}<k$, and $d$ is the channel number. We adopt a joint loss function consists of L1 loss and the Structure Similarity Index (SSIM) to simultaneously optimize the embedded feature $\mathbf{F}$ and the spatial consistency of decoded TSDF-Def tensors $\hat{\mathbf{T}}$. The loss function can be written as follows:
\begin{equation}
    \mathcal{L}
    = \lambda_{\text{L1}} \mathcal{L}_{\text{L1}}
    + \lambda_{\text{mask}} \mathcal{L}_{\text{mask}}
    + \lambda_{\text{ssim}} \mathcal{L}_{\text{ssim}},
\end{equation}
where
\begin{align}
    \mathcal{L}_{\text{L1}} &= \|\hat{\mathbf{T}} - \mathbf{T} \|_{1}, \\
    \mathcal{L}_{\text{mask}} &=\|M \cdot (\hat{\mathbf{T}} - \mathbf{T}) \|_{1}, \\
    \mathcal{L}_{\text{ssim}} &= 1 - \text{SSIM}(\hat{\mathbf{T}}, \mathbf{T}),
\end{align}
and $M = (|V[...,0]| < \alpha)$ selects the near-surface mask, where the TSDF value is below a threshold $\alpha$.

For reconstruction, the decoder $D_\psi(\cdot)$ uses 3D PixelShuffle\cite{shi2016real} between the convolution and activation layers to progressively upsample $\mathbf{F}$ back to the original TSDF-Def space. Because the decoder is also streamed, quantized 3D convolutions are applied in both the projection and decoding stages to enable model quantization to reduce model size. The decoding process is as follow:
\begin{equation}
    \hat{\mathbf{T}} = D_{\mathcal{Q}(\psi)}(\mathcal{Q}(\mathbf{F})),
\end{equation}
where $\hat{\mathbf{T}} \in \mathbb{R}^{k\times k\times k \times 4}$ is the reconstructed TSDF-Def tensors, and $\mathcal{Q}(\cdot)$ is the differentiable quantization operator. Here, while the compressed latent features capture local geometric structure, they do not, on their own, provide explicit cues about how different regions of the mesh move over time. This lack of explicit motion structure poses two key challenges for interpolation: (1) adjacent regions may deform in complex, nonlinear ways that cannot be inferred from spatial features alone, (2) without stable motion anchors, the interpolation model may produce temporally inconsistent predictions, especially across long sequences.

\begin{figure}[t]
    \centering
    \includegraphics[width=1\linewidth]{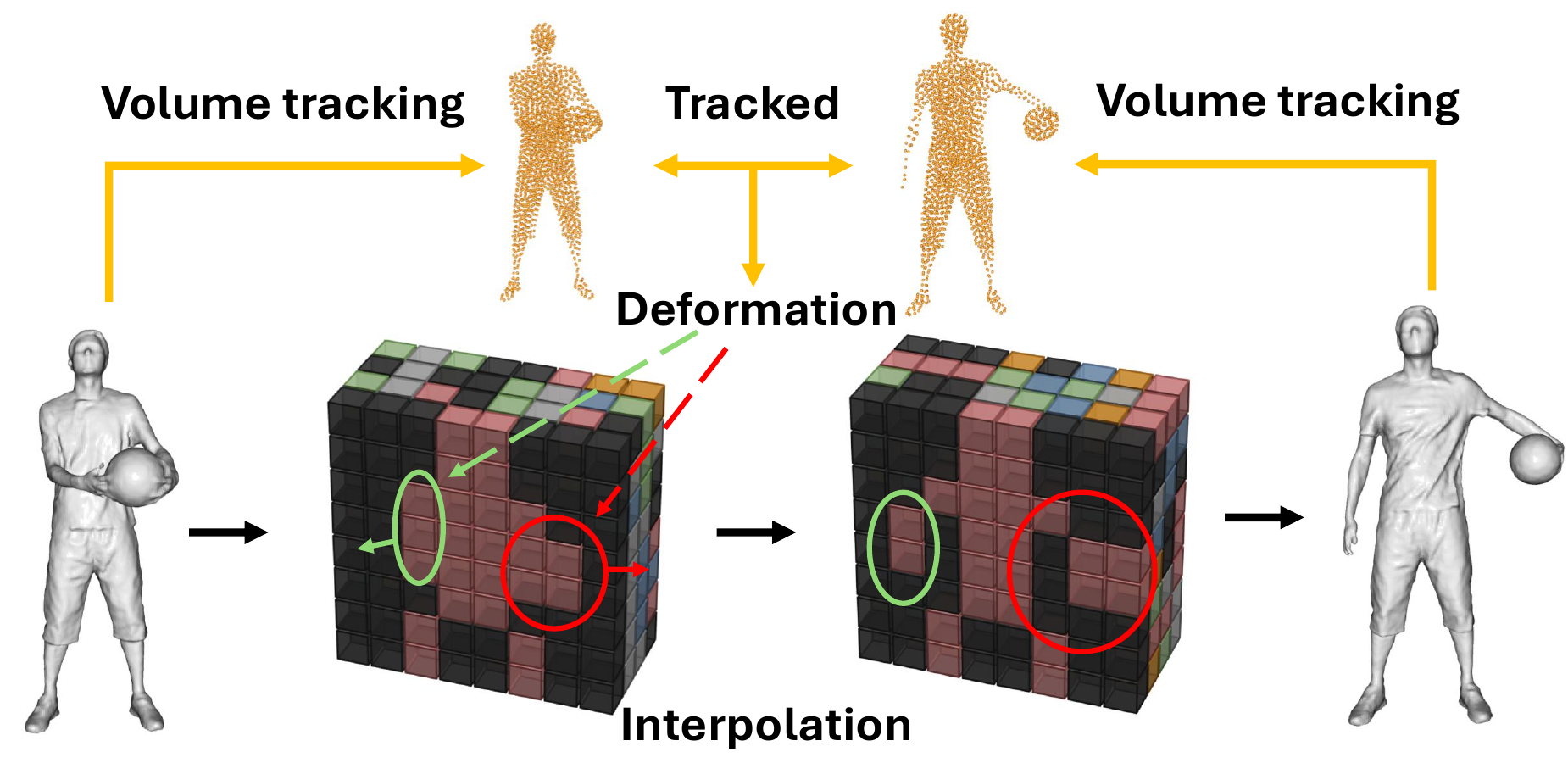} 
    \caption{Cross-section illustration of N4MC's volume tracking on input 4D mesh frames to obtain sets of volume centers (orange), and conversion of keyframe meshes (gray) into TSDF-Def tensors. For better visualization, we show the $(4\times8\times8)$ intermediate cross-sections of the TSDF tensors. N4MC leverages the tracked volume centers to extract priors for interpolation, to guide the interpolation for regions such as the hands and basketball.
    } \vspace{-0.1in}
    \label{fig:interpolation}
\end{figure}

\vspace{-0.1in}
\paragraph{Volume Tracking and Volume Centers.} To obtain meaningful motion cues that can serve as priors for interpolation, we apply volume tracking~\cite{dvovrak2022rigid, dvovrak2023global} to the input 3D mesh sequences in order to get representative volume centers  $\mathcal{C}=\{\mathbf{C}_i\}_{i=1}^{N}$, where each $\mathbf{C}_i = \{c^1, c^2, \ldots, c^p\}, c^j \in \mathbb{R}^{3}$ traces the spatial trajectory of a localized volume across all $N$ frames. Here $p$ specifies the total number of tracked centers. Because of the tracked feature of volume centers, they can capture non-rigid motion dynamics. Fig. \ref{fig:interpolation} shows the extraction of the intermediate volume center from the input meshes. Regarding how volume tracking works, we refer readers to~\cref{sec:volumetracking} in the supplementary material.

\vspace{-0.1in}
\paragraph{Volume Center-guided Motion Priors.}\label{latentcode} We then design a volume center-guided latent mapping network to embed geometric motion information into compact latent codes for transformer conditioning. We choose every $n$ frame as a key frame, and the remaining $n-1$ frames are interpolated. Specifically, given a sequence of tracked volume centers $\mathcal{C}=\{\mathbf{C}_i\}_{i=1}^{N}$, we select every $n$-th frame as a key frame and interpolate the intermediate $n-1$ frames. For each interpolation group ${\{\mathbf{C}_s, \mathbf{C}_t, \mathbf{C}_e}\}$, representing the start, target, and end frames, we encode their corresponding volume centers using a PointNet-style~\cite{qi2017pointnet} encoder $\Phi(\cdot)$ to obtain global descriptors:
\begin{equation}
    \mathbf{Z}_s = \Phi(\mathbf{C}_s), \quad \mathbf{z}_t = \Phi(\mathbf{C}_t), \quad \mathbf{z}_e = \Phi(\mathbf{C}_e).
\end{equation}
To capture non-rigid motion movements, we then calculate a temporal delta feature $\Delta \mathbf{z}_{es} = \tfrac{1}{2}(\mathbf{z}_e - \mathbf{z}_s)$. Each intermediate frame is associated with a normalized temporal index $\alpha \in [0, 1]$, which is embedded using a sinusoidal time encoder $\Gamma(\alpha)$ following~\cite{vaswani2017attention}.
The final latent representation $\mathbf{z}_{\text{latent}}$ is then obtained by concatenating the geometric and temporal features and passing them through an MLP with GELU activations and Layer Normalization:
\begin{equation}
    \mathbf{z}_{\text{latent}} = f_\omega([\mathbf{z}_s, \mathbf{z}_e, \mathbf{z}_t, \Delta\mathbf{z}_{es}, \Gamma(\alpha)]),
\end{equation}
where $f_\omega(\cdot)$ denotes MLP network. A learnable scale parameter $\sigma$ stabilizes training, and light Gaussian noise is added in training to improve robustness and exploration.

\begin{figure*}[t]
    \centering
    \includegraphics[width=1\linewidth]{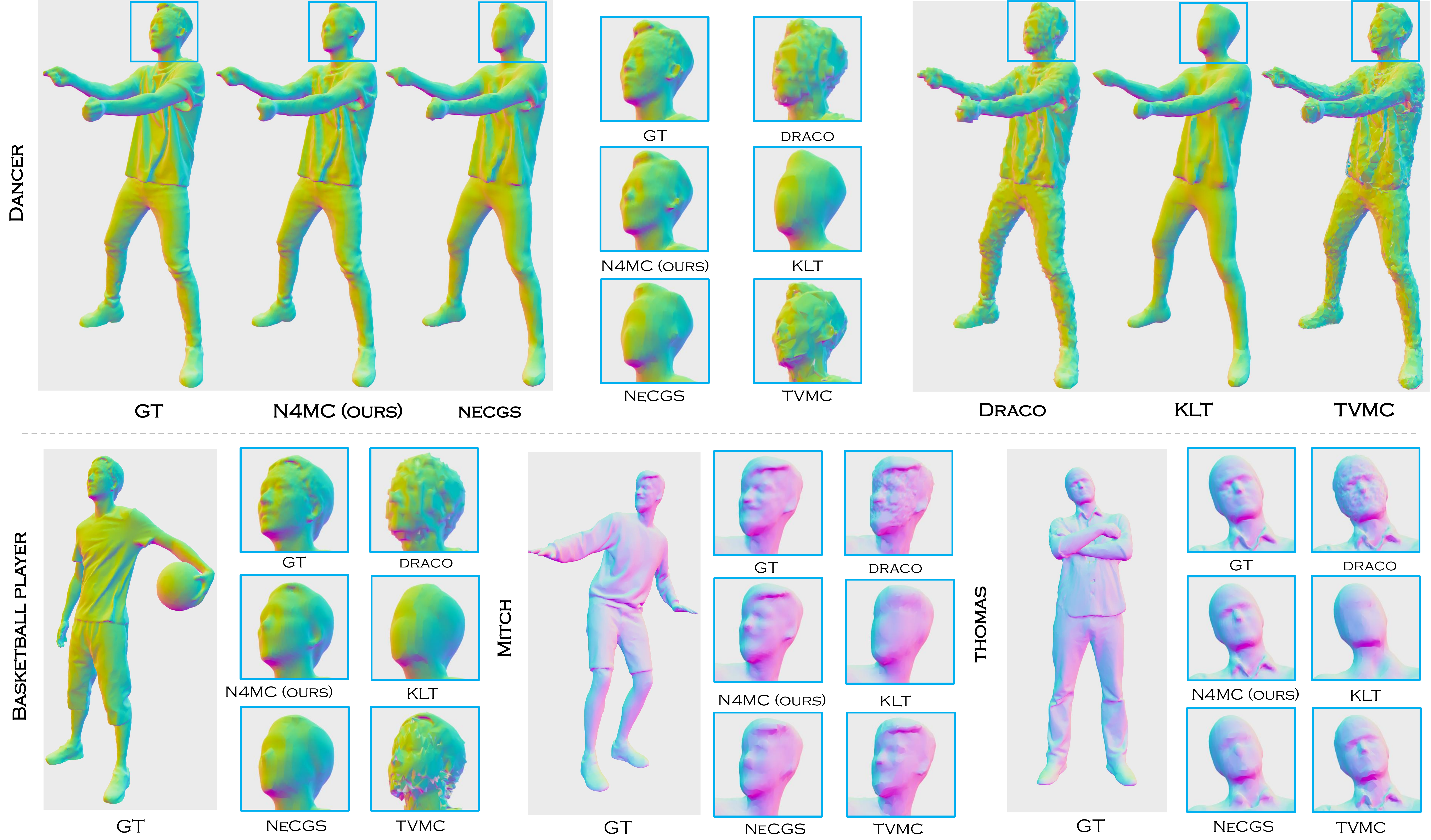} 
    \caption{Qualitative comparison of our N4MC with baselines at a bitrate around 4 Mbps. For image-based SSIM and PSNR evaluation, lighting effects are disabled, and vertex normals are converted to RGB colors before rendering. The "Dancer" and "Basketball player" sequences appear green, while "Mitch" and "Thomas" show pink due to their different orientations. We show a front-facing perspective on people here for all datasets for fair evaluation. To save space, we show full meshes for "Dancer" dataset and only ground truth with zoomed-in regions for "Basketball player", "Mitch", and "Thomas". N4MC outperforms all the baselines with very high quality.} 
    \label{fig:visual_comparison}
\end{figure*}

\subsection{3D Mesh Interpolation} 
The final stage of N4MC is the 3D mesh interpolation model. Specifically, every $n$ frame in $\mathcal{M}=\{\mathbf{M}_i\}_{i=1}^{N}$ is selected as a key frame, and the intermediate $(n-1)$ frames are interpolated via our interpolation transformer, as illustrated in~\cref{fig:interpolation}.

\vspace{-0.1in}
\paragraph{Network Architecture.}
We design a cross-attention transformer that jointly models temporal interpolation and latent-conditioned deformation synthesis. Given the embedded feature tensors of the start and end key mesh frames, $\mathbf{F}_\text{s}, \mathbf{F}_\text{e}\in\mathbb{R}^{k^\prime\times k^\prime\times k^\prime\times d}$, and the latent codes $\mathbf{Z}={\mathbf{z}_t}_{t=\text{s+1}}^{\text{e-1}}$, the network predicts intermediate embedded features $\hat{\mathbf{F}}_t$ for $t=s+1,\dots,e-1$.

We first lift each input mesh feature into a high-dimensional voxel feature space to preserve spatial detail, and then apply a linear projection to obtain a compact transformer embedding for efficient processing. Coordinate embeddings and sinusoidal time embeddings are injected to preserve spatial–temporal awareness. The encoded start–end features are concatenated and processed by a lightweight Transformer Encoder to obtain global memory tokens that serve as long-range structural context. 

For each interpolated timestep $t$, the network constructs a FiLM-conditioned query~\cite{perez2018film} that combines the latent deformation code $\mathbf{z}_t$ and time embedding. A small MLP generates a coarse prior feature from $(\mathbf{F}_\text{s}, \mathbf{F}_\text{e}, t)$, which is modulated by FiLM scaling $(\gamma_t, \beta_t)$ predicted from the latent-conditioned MLP. This prior is further refined via a cross-attention Transformer Decoder, attending to the start–end memory features to synthesize temporally consistent intermediate representations. The final output is produced through a residual prediction head and mapped back to the voxel feature space, forming $\hat{\mathbf{F}}_t \in \mathbb{R}^{k^\prime\times k^\prime\times k^\prime\times d}$.
Similar to auto-decoder design, in order to further reduce computational overhead and model size, all linear layers in the transformer are replaced with quantized linear layers. This design enables efficient inference and memory compression without significantly degrading interpolation quality.

For each interpolated timestep $t$, the interpolation process can be formally expressed as:
\begin{equation} \label{model:interpolation}
    \hat{\mathbf{F}}_t = \mathcal{I}_{\mathcal{Q}(\phi)} \mathcal{Q}[(
    \mathbf{F}_{\text{s}},
    \mathbf{F}_{\text{e}},
    \mathbf{z}_t,
    t)],
\end{equation}
where $\mathcal{I}_{\mathcal{Q}(\phi)}$ denotes the quantized cross-attention transformer parameterized by $\phi$, $\mathbf{F}_{\text{s}}$ and $\mathbf{F}_{\text{e}}$ are the voxelized features of key frames, $\mathbf{z}_t$ is the latent deformation code associated with timestep $t$, and $t \in (0,1)$ represents the normalized interpolation ratio. The model learns to synthesize temporally coherent intermediate representations $\hat{\mathbf{F}}_t$ conditioned on both structural and latent cues.

\vspace{-0.1in}
\paragraph{Loss Function.}
The interpolation transformer is trained using the same loss function as the quantization-aware auto-decoder described in ~\cref{auto-decoder}, ensuring consistency between the decoded and interpolated TSDF-Def tensors. Specifically, the loss jointly optimizes the L1, masked L1, and SSIM terms between the decoded TSDF-Def tensors of the interpolated frames and their ground truth counterparts. This ensures that the interpolated embeddings not only preserve geometric fidelity but also maintain temporal and structural consistency across the entire mesh sequence.

\subsection{Overall Compression and Decompression}
During compression, each frame $\mathbf{M}_i$ is first converted into its corresponding TSDF-Def tensor $\mathbf{T}_i$ using the TSDF-Def generation module described in ~\cref{auto-decoder}. $\mathbf{Z}_{\text{latent}}$ is then generated as described in~\cref{latentcode}. The encoder $E_\theta(\cdot)$ then transforms $\mathbf{T}_i$ into compact embedded features $\mathbf{F}_i = E_\theta(\mathbf{T}_i)$, which are subsequently quantized through $\mathcal{Q}(\cdot)$ to obtain $\mathcal{Q}(\mathbf{F}_i)$. 
Only the embedded features of key frames and intermediate corresponding latent codes $\mathbf{z}_{\text{latent}}$ are stored. Entropy coding is applied to further compress the embedded features, latent codes and sequence-specific models into bitstreams. We adopt Huffman codec~\cite{huffman2007method} but other advanced coding methods can be used.

At decompression time, the decoder reconstructs the full mesh sequence from the compressed embedded features $\mathcal{F}$ and latent codes $\mathbf{Z}$. The quantized transformer $\mathcal{I}_{\mathcal{Q}(\phi)}$ interpolates the intermediate embedded features $\hat{\mathbf{F}}_t$ from key frame embeddings and latent deformation codes using Equation~\ref{model:interpolation}.
The reconstructed embedded features are then passed through the quantized auto-decoder $D_{\mathcal{Q}(\psi)}(\cdot)$ to recover the full-resolution TSDF-Def tensors:
\begin{equation}
    \hat{\mathbf{T}}_t = D_{\mathcal{Q}(\psi)}(\hat{\mathbf{F}}_t).
\end{equation}
Finally, the Deformable Marching Cubes algorithm~\cite{ren2024necgs} extracts the corresponding 3D mesh $\hat{\mathbf{M}}_t$ from $\hat{\mathbf{T}}_t$:
\begin{equation}
    \hat{\mathbf{M}}_t = \text{DMC}(\hat{\mathbf{T}}_t), \quad t = 1, \dots, N.
\end{equation}

\begin{table*}[t]
\centering
\setlength{\tabcolsep}{3pt}
\renewcommand{\arraystretch}{1.2}
\caption{Quantitative comparison with N4MC and baselines in terms of decoding time, D2-PSNR, image-based SSIM, and PSNR with a similar bitrate around 4 Mbps. The best results are shown in orange, and the second-best in yellow. Corresponding qualitative (visual) comparisons are shown in~\cref{fig:visual_comparison}. 
}
\resizebox{\textwidth}{!}{
\begin{tabular}{lccccccccccccccccc}
\toprule
& &
\multicolumn{4}{c}{\textbf{Dancer}} & \multicolumn{4}{c}{\textbf{Basketball Player}} 
& \multicolumn{4}{c}{\textbf{Mitch}} & \multicolumn{4}{c}{\textbf{Thomas}} \\

\cmidrule(lr){3-6}\cmidrule(lr){7-10}\cmidrule(lr){11-14}\cmidrule(lr){15-18}
Method & Time [ms]\,$\downarrow$ & D2-PSNR$\uparrow$ & SSIM$\uparrow$ & PSNR$\uparrow$ & Bitrate  & D2-PSNR$\uparrow$ & SSIM$\uparrow$ & PSNR$\uparrow$ & Bitrate  &
 D2-PSNR$\uparrow$ & SSIM$\uparrow$ & PSNR$\uparrow$ & Bitrate  & D2-PSNR$\uparrow$ & SSIM$\uparrow$ & PSNR$\uparrow$ & Bitrate   \\
\midrule

Draco~\cite{Draco2024} & \cellcolor{orange!30} 2.22 & 54.308 & 0.913 & 24.08 & 3.986 & 57.577 & 0.904 & 23.755 & 4.046 & 63.696 & 0.948 & 27.1 & 4.108 & 63.677 & 0.952 & 28.07 & 4.161\\
KLT~\cite{realtime2018} & 24.40 & 58.166 & \cellcolor{yellow!30}0.9490 & 28.027 & 4.064 & 58.511 &  0.9431 & 27.792 & 4.036  & 62.469 & 0.9450 & 28.344 & 4.091 & 62.026 & 0.9518 & 29.494 & 3.870  \\
TVMC~\cite{chen2025tvmc} &\cellcolor{yellow!30}  11.29 & \cellcolor{yellow!30} 65.527 & 0.9388 &\cellcolor{yellow!30} 30.2838 & 4.237 &  \cellcolor{yellow!30}61.840 & 0.8795 & 25.889 & 4.438 & \cellcolor{yellow!30} 72.807 & 0.9639 & \cellcolor{yellow!30}34.37 & 4.664 &\cellcolor{yellow!30} 71.064 & 0.9639 & \cellcolor{yellow!30}34.328 & 4.195\\
NeCGS~\cite{ren2024necgs} & 37.39 & 62.065 & 0.9458 & 26.405 & 3.480 & 61.228 & \cellcolor{yellow!30}0.9540 & \cellcolor{yellow!30}29.764 & 3.656 & 69.256 & \cellcolor{yellow!30}0.9647 & 32.337 & 4.298  & 67.157 & \cellcolor{yellow!30}0.9670 & 33.051 & 3.520\\

\textbf{N4MC (Ours)} & \textbf{38.63} & \cellcolor{orange!30} \textbf{67.276} & \cellcolor{orange!30} \textbf{0.9712} & \cellcolor{orange!30} \textbf{33.567} & \textbf{4.514} & \cellcolor{orange!30} \textbf{66.107} & \cellcolor{orange!30} \textbf{0.9671} &
\cellcolor{orange!30} \textbf{33.188} & \textbf{4.311} & \cellcolor{orange!30} \textbf{73.129} & \cellcolor{orange!30} \textbf{0.9779} & \cellcolor{orange!30} \textbf{37.274} & \textbf{4.430} & \cellcolor{orange!30} \textbf{72.759} & \cellcolor{orange!30} \textbf{0.9791} & \cellcolor{orange!30} \textbf{37.811} & \textbf{4.269} \\

\bottomrule
\end{tabular}
}
\label{tab:comparison}
\end{table*}

\section{Experiments}\label{experiments}

\paragraph{Datasets.} We evaluate N4MC on a diverse suite of 4D mesh sequences covering real, synthetic, single-object, and multi-object scenarios. Our primary benchmarks include four complex time-varying mesh sequences from the MPEG V-DMC standard~\cite{isoCfp}: "Dancer", "Basketball player", "Mitch", and "Thomas", each with 300 frames~\cite{yang2023tdmd}. In addition, we construct a "Mixed" dataset by synchronizing and combining 4 of these sequences to further evaluate N4MC's scalability and robustness in complex multi-object scenarios. We also evaluate N4MC on two additional sources: (1) our own custom captured sequences featuring everyday human activities and object interactions, and (2) high-variability synthetically generated data from Thingi10K~\cite{Thingi10K}. Results on these additional datasets are provided in supplementary material in~\cref{sec:addresults}.

\vspace{-0.1in}
\paragraph{Baselines.}
We compare N4MC with several recent mesh compression methods/tools, including both static mesh compression and classical algorithmic time-varying mesh compression, explicitly and implicitly, for a comprehensive comparison.
We use the official implementation for those methods with source code releases for and use a re-implemented version otherwise. Specifically, we compare against NeCGS~\cite{ren2024necgs}, TVMC~\cite{chen2025tvmc}, Draco~\cite{Draco2024}, and Google KLT~\cite{realtime2018}

\vspace{-0.1in}
\paragraph{Metrics.}
We evaluate the quality of decoded meshes using both the image-based and geometry-based Peak Signal-to-Noise Ratio (PSNR) and Structural Similarity Index Measure (SSIM) metrics. Following the MPEG standard~\cite{3DG2019metrics}, D2-PSNR is widely used in mesh compression to assess geometric accuracy, 
but for time-varying meshes without explicit correspondence, nearest-neighbor matching is typically adopted, which limits the reliability of PSNR in reflecting true perceptual quality. 
To complement this, we compute image-based PNSR and SSIM metrics from four viewpoints. To enhance the accuracy and perceptual relevance of SSIM, we disable lighting effects and convert vertex normals into RGB colors before rendering, ensuring consistent visual comparison across frames.

\begin{figure}[t]
     \centering
     \begin{subfigure}[b]{0.22\textwidth}
         \centering
         \includegraphics[width=1\textwidth]{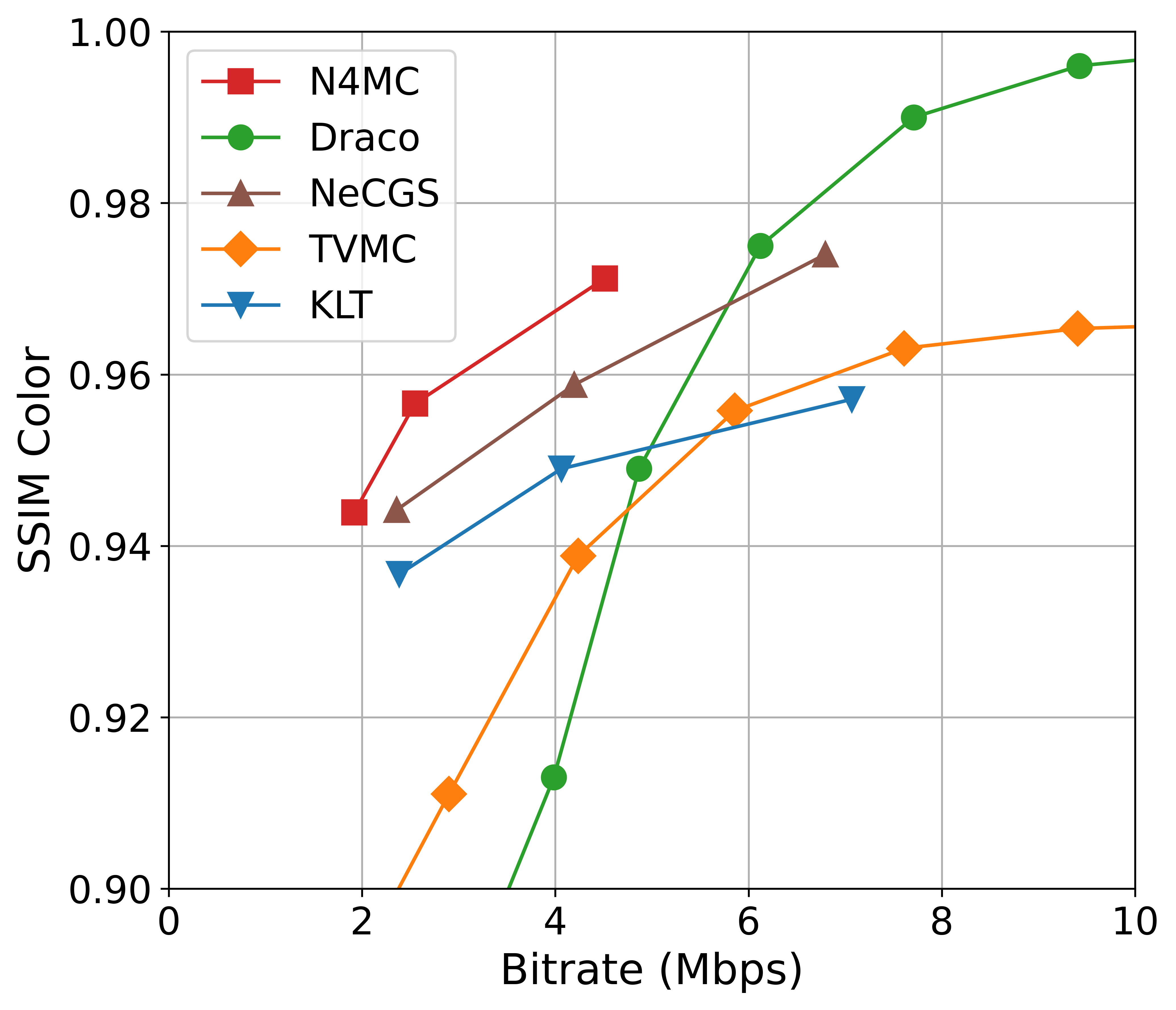}
         \caption{RD performance and comparison on "Dancer" mesh.}
         \label{fig:compare:a}
     \end{subfigure}
     \hfill
     \begin{subfigure}[b]{0.22\textwidth}
         \centering
         \includegraphics[width=1\textwidth]{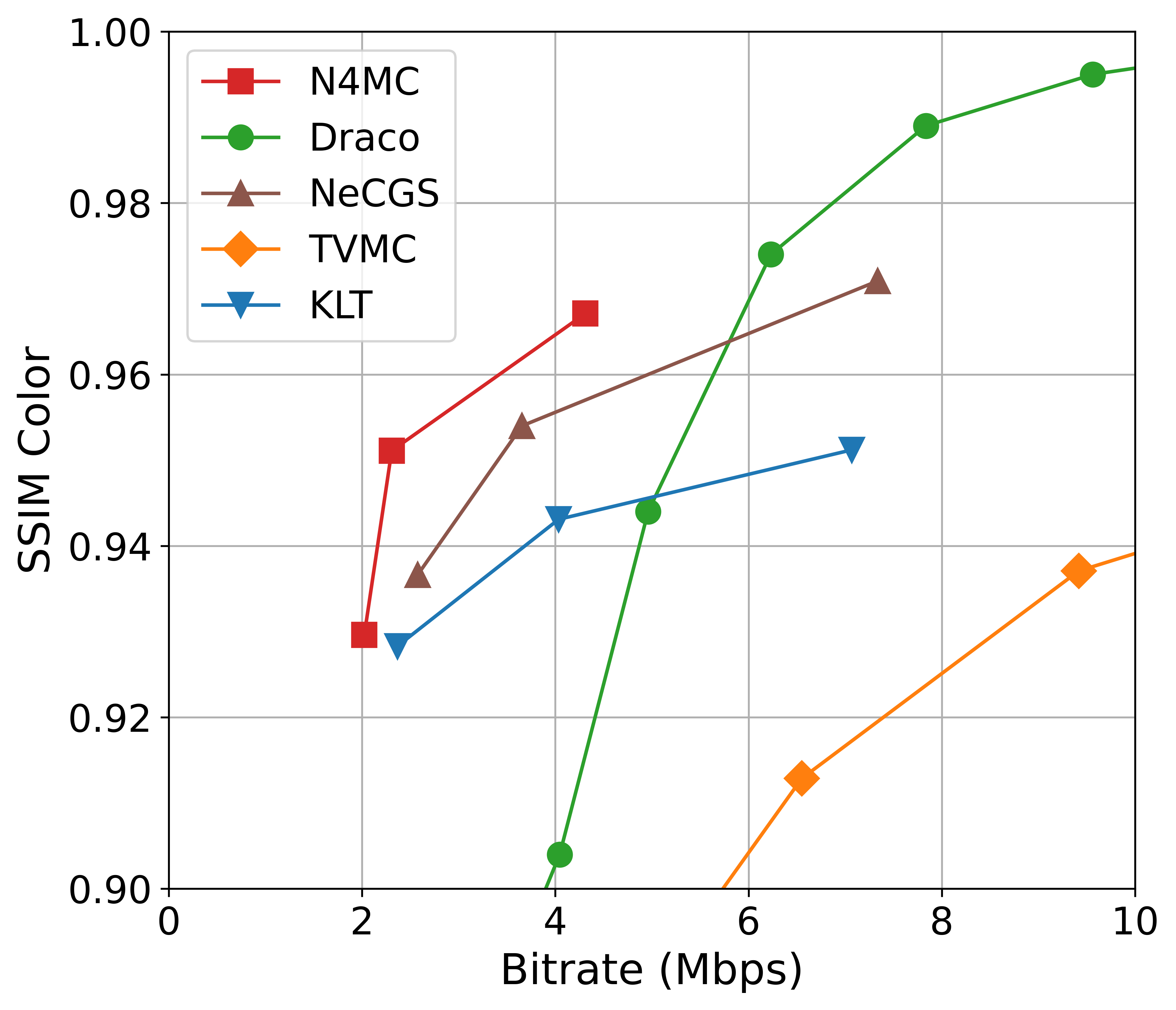}
         \caption{RD performance and comparison on "Basketball player" mesh.}
         \label{fig:compare:b}
     \end{subfigure}
     \hfill
     \begin{subfigure}[b]{0.22\textwidth}
         \centering
         \includegraphics[width=1\textwidth]{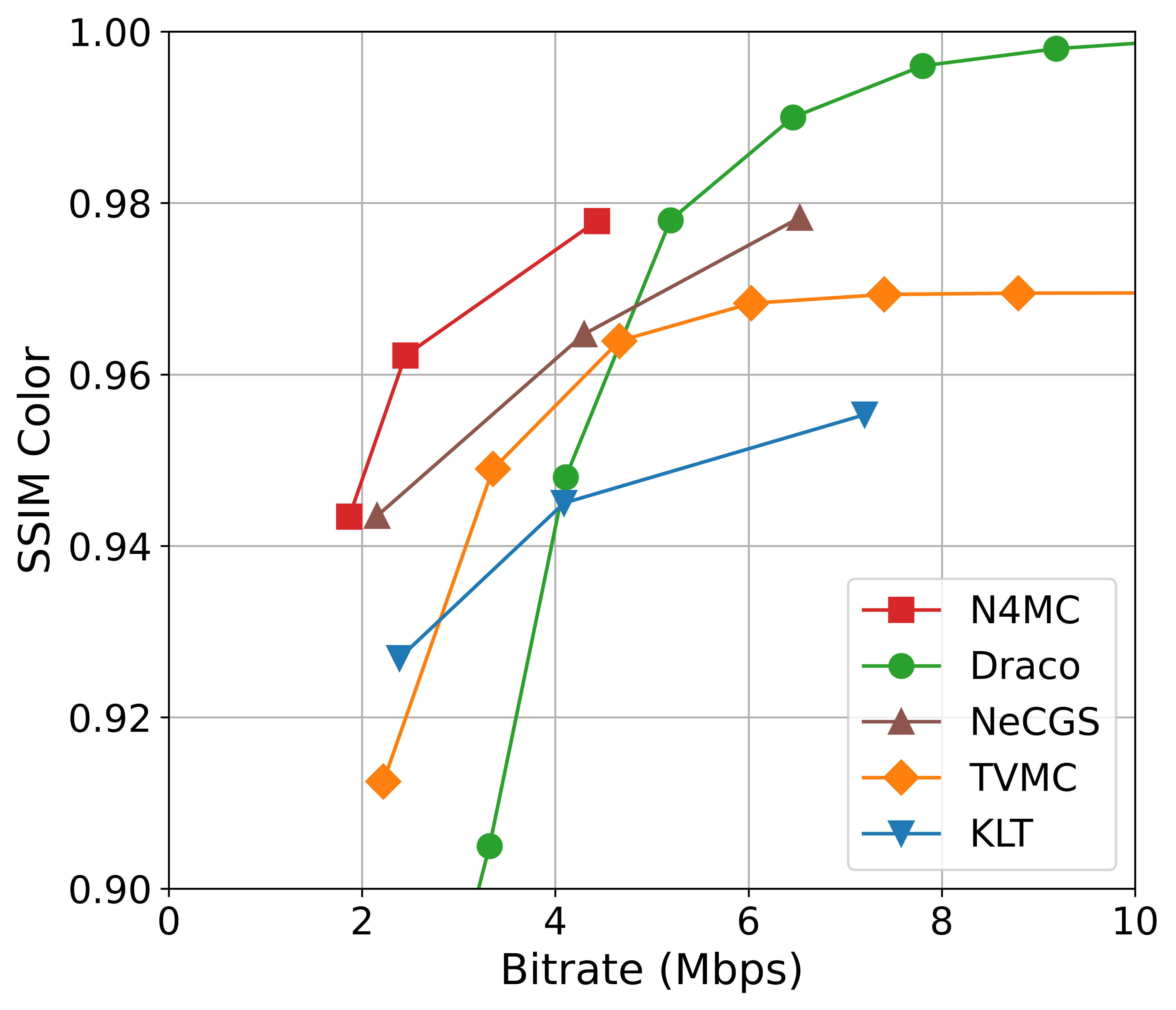}
         \caption{RD performance and comparison on "Mitch" mesh.}
         \label{fig:compare:c}
     \end{subfigure}
     \hfill
     \begin{subfigure}[b]{0.22\textwidth}
         \centering
         \includegraphics[width=1\textwidth]{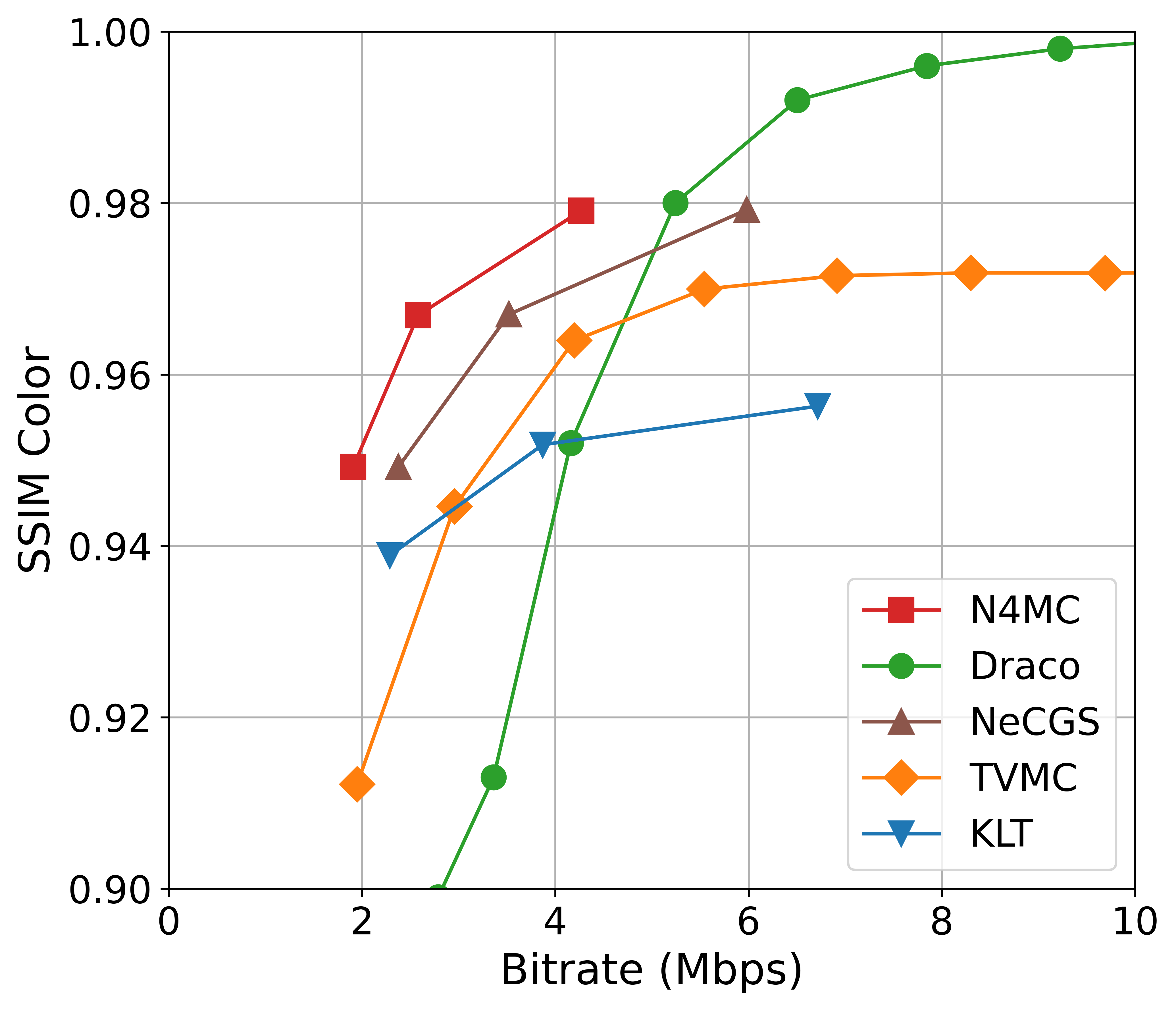}
         \caption{RD performance and comparison on "Thomas" mesh.}
         \label{fig:compare:d}
     \end{subfigure}
        \caption{Objective rate–distortion (RD) performance comparison of image-based SSIM versus bitrate on MPEG sequences. To get the target bitrates, for TVMC~\cite{chen2025tvmc} and Draco~\cite{Draco2024}, the quantization parameter $qp$ is varied from 4 to 14. For NeCGS~\cite{ren2024necgs} and our N4MC, TSDF-Def resolutions of 64, 128, and 256 are used. For KLT~\cite{realtime2018}, 16, 32, and 128 basis vectors are employed.}
        \label{fig:compare}
        \vspace{-0.2in}
    \label{RD-performance}
\end{figure}

\vspace{-0.1in}
\paragraph{Results.}
In \cref{RD-performance}, we compare N4MC with all baselines in terms of SSIM versus Bitrates at an evaluation frame rate of 30 FPS. For N4MC and NeCGS~\cite{ren2024necgs}, we vary the TSDF-Def resolution among 64, 128, 256, adjusting the embedded feature dimensions to $(4,4,4,64)$, $(8,8,8,16)$, and $(8,8,8,32)$, respectively. We further control the \textit{transformer dimension} $\Lambda$ with values of 16, 24, and 32 to balance model capacity and size. The larger $\Lambda$, the more expressive but heavier the model becomes. For TVMC~\cite{chen2025tvmc} and Draco~\cite{Draco2024}, we vary the quantization parameter $qp$ between 4 to 14 to generate rate-distortion curves.

As shown in \cref{RD-performance}, N4MC consistently outperforms these baselines. It achieves higher compression efficiency than NeCGS~\cite{ren2024necgs} while achieving comparable quality with the help of 3D mesh interpolation. Compared with TVMC~\cite{chen2025tvmc} and the static mesh compressor Draco~\cite{Draco2024}, N4MC delivers substantially higher visual quality at similar bitrates (around 4 Mbps). For the "Basketball Player" dataset, we observed a noticeable quality drop in TVMC~\cite{chen2025tvmc}. This is because the Basketball Player dataset exhibits larger motion variations compared with others, leading to greater distortions for deformation-based methods such as TVMC~\cite{chen2025tvmc}, particularly in terms of the image-based metric SSIM. Nevertheless, it still maintains a relatively high D2-PSNR, as shown in \cref{tab:comparison}. This observation further justifies our use of both D2-PSNR and image-based SSIM and PSNR for a more comprehensive evaluation.

To further evaluate reconstruction fidelity, we present qualitative results for four datasets at approximately 4 Mbps in ~\cref{fig:visual_comparison}. As shown, N4MC produces noticeably fewer distortions and better preserves fine geometric details such as facial expressions and thin structures. Quantitative comparisons summarized in \cref{tab:comparison}, which shows decoding time, D2-PSNR, SSIM, PSNR, and bitrate, demonstrate that, under similar bitrate conditions, N4MC achieves the highest reconstruction quality among all compared methods. 

We also compare the per-frame decoding time of N4MC with all baselines on RTX 4090 GPU. As shown in \cref{tab:comparison}, Draco~\cite{Draco2024} achieves the fastest decoding speed, requiring only about 2 ms per frame for single human objects. In contrast, both N4MC and NeCGS~\cite{ren2024necgs} take longer due to the processes of constructing voxel grids and applying the Marching Cubes algorithm~\cite{lorensen1998marching}. Nonetheless, this still shows a real-time decoding performance of over 24 FPS. More visual results are reported in~\cref{sec:addresults} in the supplementary material and in \textbf{supplementary video}.

\begin{table}[t]
\centering
\setlength{\tabcolsep}{8pt}
\renewcommand{\arraystretch}{1.2}
\caption{Ablation study at different resolutions, evaluated in terms of D2-PNSR, image-based SSIM, PSNR, and decoding time.}
\resizebox{\linewidth}{!}{
\begin{tabular}{lccccc}
\toprule
\textbf{Resolution ($k$)} & D2-PSNR$\uparrow$ & SSIM$\uparrow$ & PSNR$\uparrow$ & Bitrate & Time [ms]\,$\downarrow$ \\
\midrule
64 & 54.986 & 0.9296 & 24.721 & 2.024 & 4.097\\
128 & 59.633 & 0.9511 & 29.278 & 2.308 & 7.371\\
256 & 66.107 & 0.9671 & 33.188 & 4.311 & 38.537\\
\bottomrule
\end{tabular}
}
\label{tab:ablation_resolution}
\end{table}

\begin{table}[t]
\centering
\setlength{\tabcolsep}{8pt}
\renewcommand{\arraystretch}{1.2}
\caption{Ablation on \textbf{group size} and \textbf{total frame number}.}
\resizebox{\linewidth}{!}{
\begin{tabular}{lcccccc}
\toprule
\textbf{$n$}& $N$ & D2-PSNR$\uparrow$ & SSIM$\uparrow$ & PSNR$\uparrow$ & Bitrate & Time [ms]\,$\downarrow$ \\
\midrule
5  & 100 & 68.225 & 0.9622 & 31.847 & 2.442 & 6.935 \\
10 & 100 & 67.632 & 0.9610 & 31.494 & 2.071 & 6.417\\
15 & 100 & 67.040 & 0.9597 & 31.129 & 1.968 & 6.227\\
5 & 200 & 67.635 & 0.9608 & 31.480 & 2.283 & 7.112\\
5 & 300 & 64.484 & 0.9517 & 29.521 & 1.842 & 6.868\\
\bottomrule
\end{tabular}
}
\label{tab:ablation_group_size}
\end{table}

\begin{figure}
     \centering
     \begin{subfigure}[b]{0.11\textwidth}
         \centering
         \includegraphics[width=\textwidth]{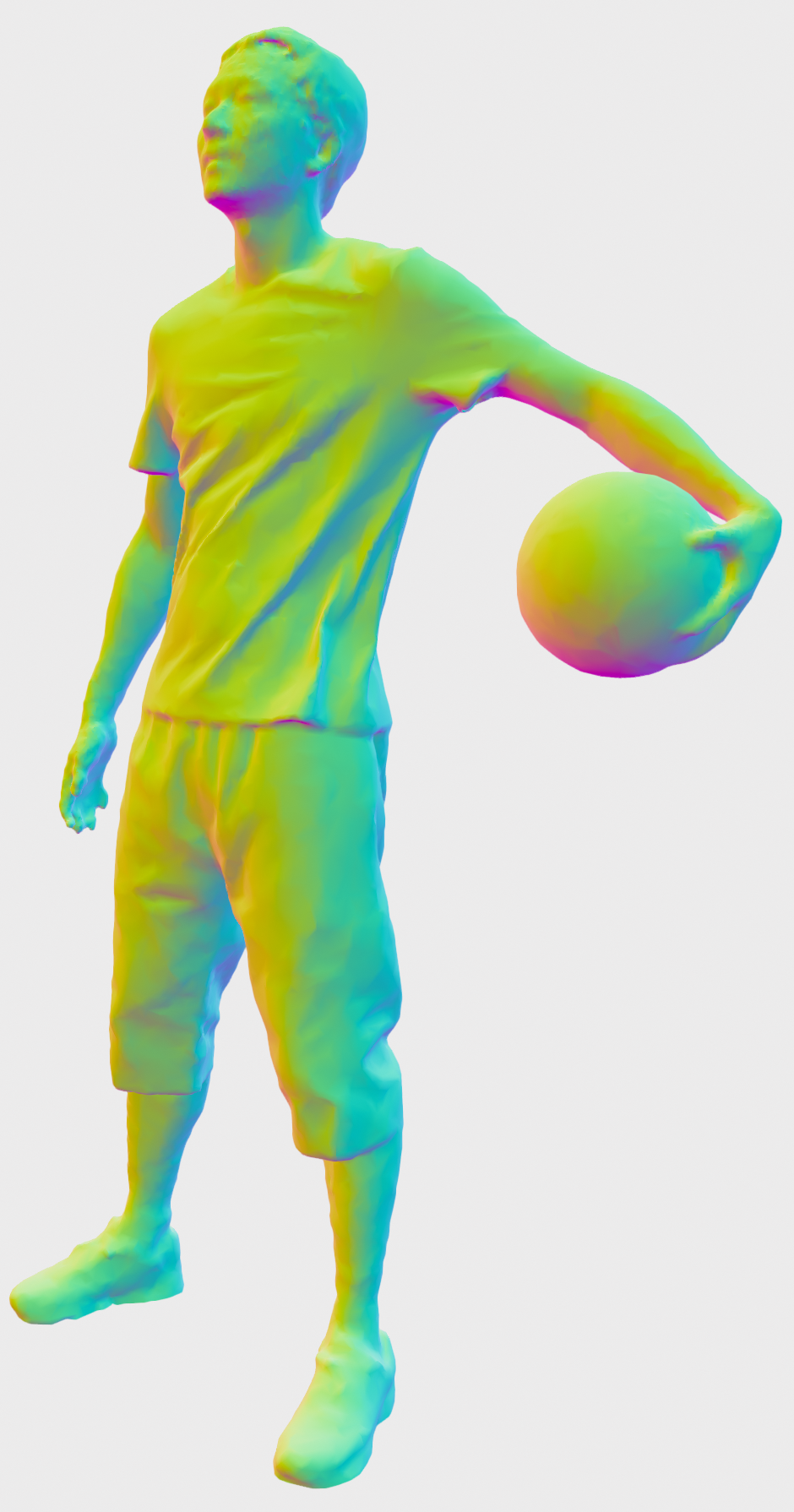}
         \caption{gt}
         \label{1000centers}
     \end{subfigure}
     \hfill
     \begin{subfigure}[b]{0.11\textwidth}
         \centering
         \includegraphics[width=\textwidth]{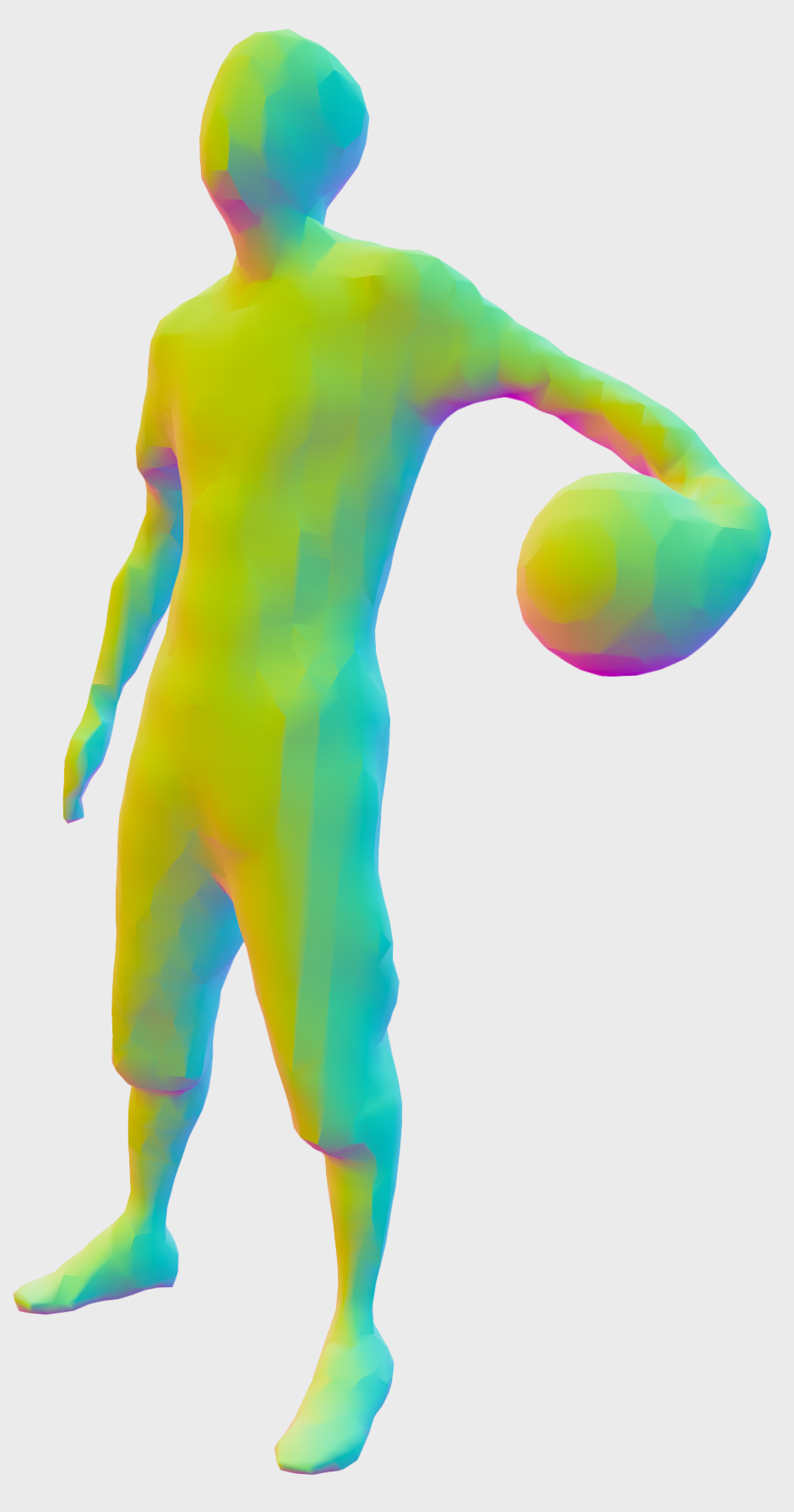}
         \caption{64}
         \label{2000centers}
     \end{subfigure}
     \hfill
     \begin{subfigure}[b]{0.11\textwidth}
         \centering
         \includegraphics[width=\textwidth]{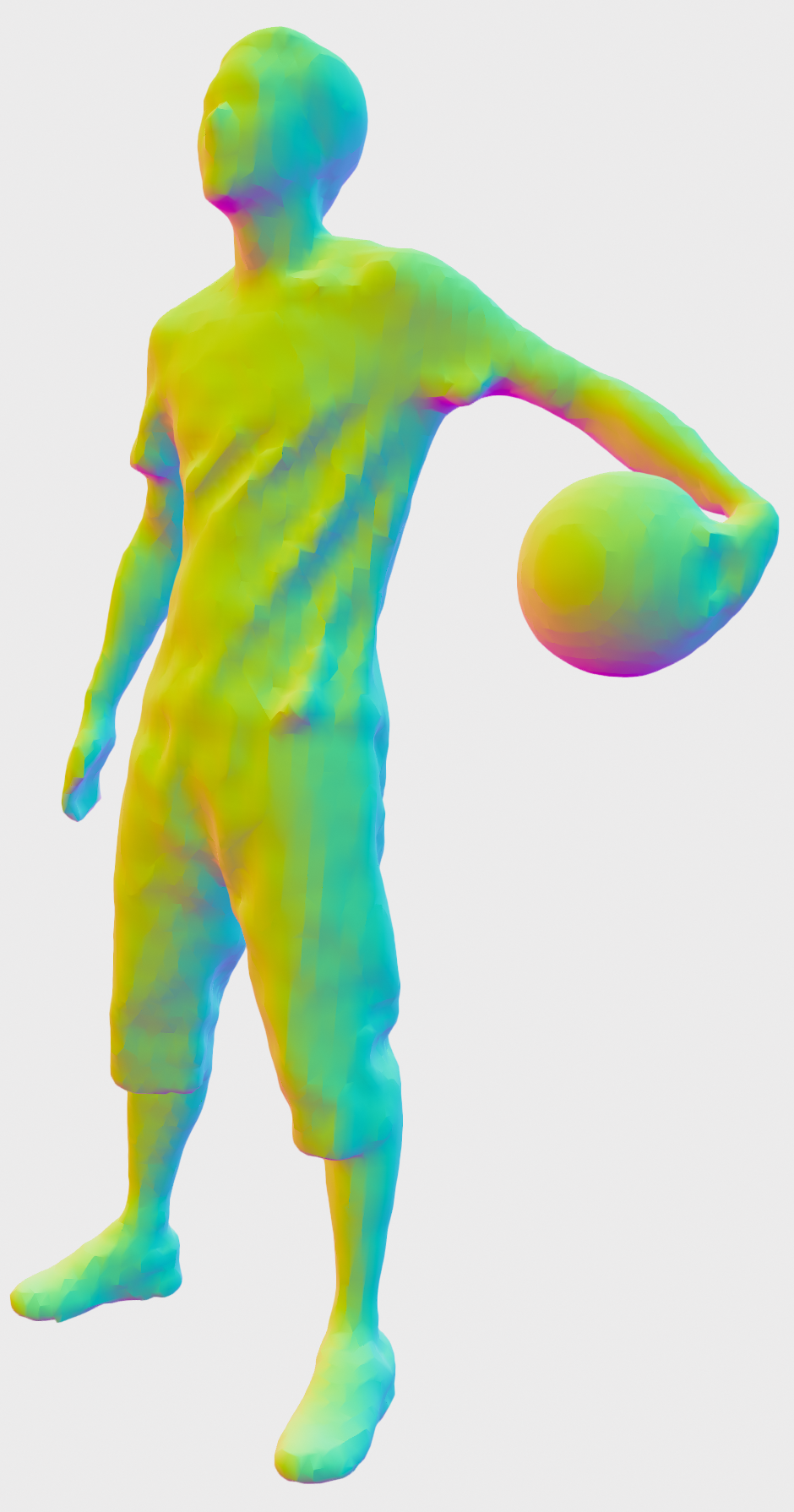}
         \caption{128}
         \label{2000centers}
     \end{subfigure}
     \hfill
     \begin{subfigure}[b]{0.11\textwidth}
         \centering
         \includegraphics[width=\textwidth]{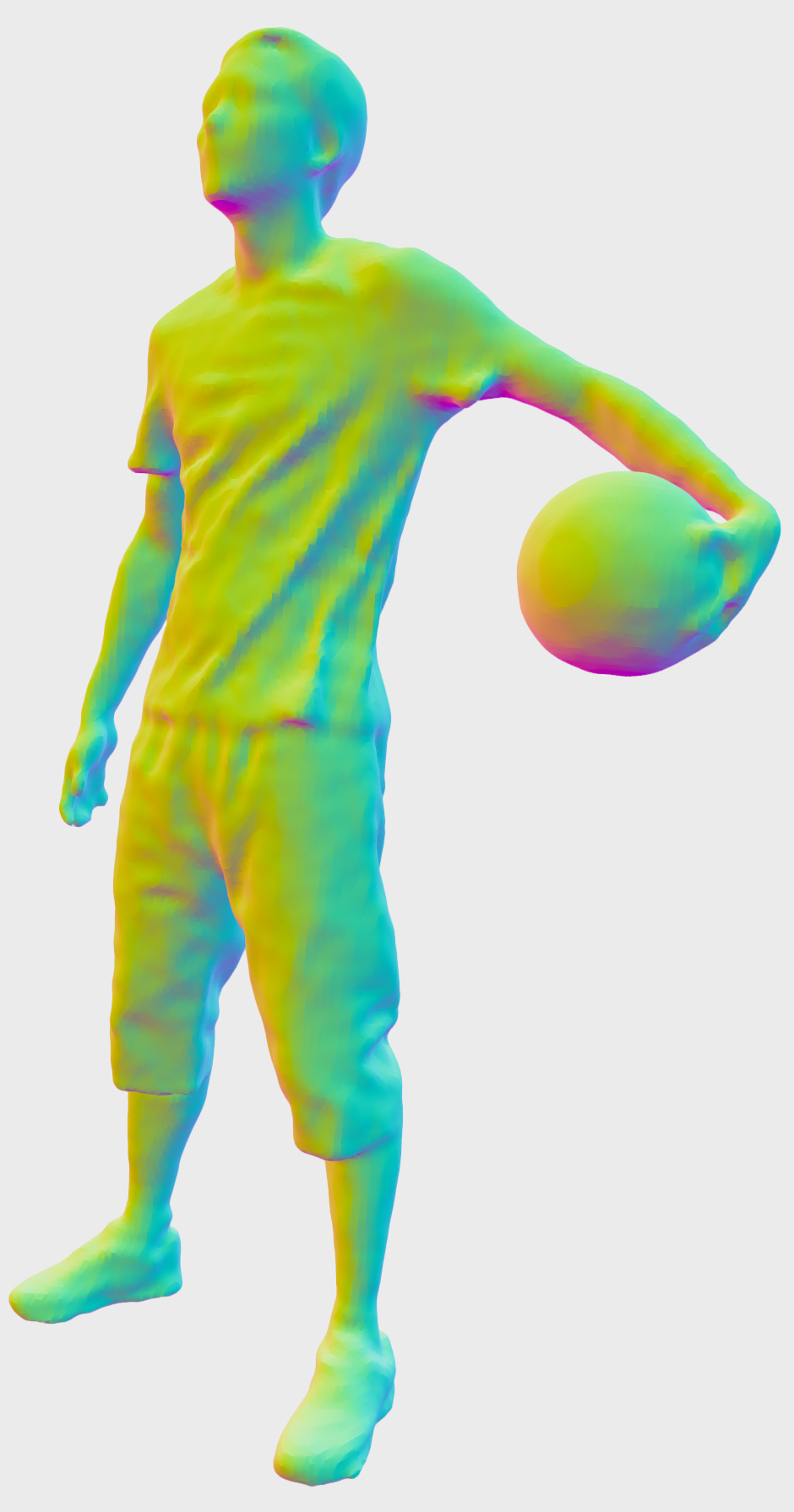}
         \caption{256}
         \label{2000centers}
     \end{subfigure}
        \caption{Qualitative comparisons of N4MC with different TSDF-Def resolutions of 64, 128, and 256.}
        \label{fig:differentresolutions}
\end{figure}

\vspace{-0.1in}
\paragraph{Ablations.} In \cref{RD-performance}, we present rate-distortion results of N4MC with the resolution $k$ of 64, 128, and 256 with the auto-decoder structure adjusted accordingly. Lower resolutions adopt a simpler architecture, and vice versa. The quantitative and qualitative comparisons of N4MC on "Dancer" dataset are shown in \cref{tab:ablation_resolution} and \cref{fig:differentresolutions}, respectively. Increasing the TSDF-Def resolution effectively enhances the quality of decoded meshes, preserving finer structures details after decoding, as shown in \cref{fig:differentresolutions}. However, it also leads to longer decoding times and higher bitrates.

We also conduct experiments to investigate how the group size $n$ and the total number of frames $N$ affect decoding performance. As shown in \cref{tab:ablation_group_size}, increasing the group size or the total number of frames leads to a decrease in decoding quality. A larger group size $n$ slightly reduces the interpolation time per frame, thereby lowering the decoding time per frame. In contrast, increasing the total number of frames has no significant impact on decoding time.

To further evaluate the scalability and robustness of N4MC in complex multi-object scenarios, we compare it with all baselines on the “Mixed” dataset, which consists of four 4D mesh sequences. As shown in \cref{tab:comparison_mixed}, to achieve a similar quality, Draco~\cite{Draco2024} and TVMC~\cite{chen2025tvmc} exhibit significantly higher bitrate, as their performance heavily depends on the geometric properties of the input meshes, such as the number of vertices. In contrast, N4MC maintains high reconstruction quality with relatively low bitrate requirements. This is because, although the “Mixed” dataset features more geometrically complex content, it is converted into TSDF tensors with the same shape, which N4MC can process robustly and efficiently.

 \vspace{-0.1in}

\begin{table}[t]
\centering
\setlength{\tabcolsep}{5pt}
\renewcommand{\arraystretch}{1.2}
\caption{Quantitative comparison on the \textbf{Mixed} dataset at similar quality levels to illustrate bitrate differences.}
\resizebox{0.9\linewidth}{!}{
\begin{tabular}{lccccc}
\toprule
Method & Time [ms]\,$\downarrow$ & D2-PSNR$\uparrow$ & SSIM$\uparrow$ & PSNR$\uparrow$ & Bitrate \\
\midrule
Draco~\cite{Draco2024}  & 9.10 & 57.597 & 0.899 & 27.437 & 14.487 \\
KLT~\cite{realtime2018} & 24.7 & 61.450 & 0.9424 & 30.479 & 10.737 \\
TVMC~\cite{chen2025tvmc} & 36.457 & 60.385 & 0.8820 & 27.366 & 9.847 \\
NeCGS~\cite{ren2024necgs} & 6.89 & 61.739 & 0.9440 & 30.555 & 4.931 \\
\textbf{N4MC (Ours)} & \textbf{7.33} & \textbf{61.121} & \textbf{0.9399} & \textbf{29.898} &\textbf{2.469}\\
\bottomrule
\end{tabular}
}
\label{tab:comparison_mixed}
\end{table}

\vspace{-0.1in}
\paragraph{N4MC Decoder on Mobile Devices.}
We evaluate the decoding performance of our low-resolution N4MC model on three mobile devices, Meta Quest 3, Pixel 9, and Xiaomi 12S Pro (\cref{tab:compare_mobile_time}). The decoding time consists of two parts: model inference, which reconstructs 4D tensors from the embedded features, and Marching Cubes algorithm, which extract meshes from tensors for playback. To the best of our knowledge, N4MC is the first neural 4D mesh decoder capable of running entirely on mobile devices, including a standalone VR headset (Quest 3). Due to the hardware limitations of mobile devices, all three devices exhibit noticeably longer overall decoding times.
\begin{table}[h]
\centering
\caption{Decoding time comparison across mobile devices.}
\resizebox{1\linewidth}{!}{
\begin{tabular}{lccc}
\toprule
 & Meta Quest 3 & Pixel 9 & Xiaomi 12S Pro \\
\midrule
Overall time [ms] & 346.02 & 144.34 & 156.16 \\
Model Inference [ms]   & 235.29 & 96.28 & 100.95 \\
Marching Cubes [ms] & 110.73 & 48.06 & 55.21 \\
\bottomrule
\end{tabular}
}
\label{tab:compare_mobile_time}
\end{table}
\vspace{-0.2in}

\section{Conclusion}
We present N4MC, the first neural 4D mesh compression framework. We introduce sequence-specific and lightweight models that enable real-time decoding for 4D mesh sequences. N4MC also provides a Unity plugin for mobile devices (e.g., Meta Quest 3) decoding and playback. In our extensive evaluation covering real, synthetic, single-object, and multi-object scenarios, we demonstrate that N4MC outperforms existing state-of-the-art baselines in both quantitative and qualitative evaluations, while maintaining a relatively fast decoding. See supplementary material for more details of N4MC and extensive evaluations.
{
    \small
    \bibliographystyle{ieeenat_fullname}
    \bibliography{reference}
}
\clearpage
\setcounter{page}{1}
\maketitlesupplementary

In this supplementary material, we complete the detailed introduction of volume tracking in~\cref{sec:volumetracking}, provide additional implementation details in~\cref{sec:model} and show more evaluation results in~\cref{sec:addresults}. Code and demo video are also provided.

\section{Volume Tracking}
\label{sec:volumetracking}
Given a time-varying mesh sequence $\mathcal{M}=\{\mathbf{M}_i\}_{i=1}^{N}$, to encode long-term structural consistency, we generate a set of volume centers $\mathcal{C}=\{\mathbf{C}_i\}_{i=1}^{N}$, where each $\mathbf{C}_i = \{c^1, c^2, \ldots, c^p\}, c^j \in \mathbb{R}^{3}$ traces the spatial trajectory of a localized volume across all $N$ frames. Here $p$ specifies the total number of tracked centers, in this paper we set $p=2000$ for all experiments. Each center represents a small volume surrounding it whose position evolves over time. To track the volume centers, we adopt As-Rigid-As-Possible Volume Tracking and set the mode to \textit{max affinity based tracking}, as proposed in \cite{dvovrak2023global}, to the time-varying mesh sequence~$\mathcal{M}=\{\mathbf{M}_i\}_{i=1}^{N}$.

To generate uniformly distributed centers, which ensure consistent spatial coverage of the mesh and enable stable motion tracking, we first set a resolution based on the longest axis of the mesh and convert each frame into a tightly bounded regular voxel grid. We then compute the fast winding number~\cite{barill2018fast} to define an indicator function (IF), which identifies the voxels enclosed by the mesh as follows:
\begin{equation}
    IF(x) = \left\{
    \begin{aligned}
        1 & & \omega_{f}(x) > \mu \\
        0 & & \omega_{f}(x) \leq \mu
    \end{aligned}
    \right.,
\end{equation}
where $\omega_{f}(x)$ represents the fast winding number calculated from point $x$, and $\mu$ is a threshold. In this paper, we set $\mu=0.5$. The algorithm then samples $p$ random voxels with $IF(x)=1$ and designates them as centers. Each of the remaining voxels is then connected with its nearest center, and every center is moved to its surrounding neighbors' centroid iteratively. This process yields a uniformly distributed set of centers in the first frame. For subsequent frames, the previous frame’s centers are linearly extrapolated, followed by energy-based optimization to maintain uniformity and temporal smoothness, following the approach in~\cite{dvovrak2022rigid, dvovrak2023global}.

\section{Implementation Details}
\label{sec:model}
\paragraph{Auto-encoder-decoder Pair}
For the auto-encoder, it is built upon a lightweight 3D ConvNeXt~\cite{liu2022convnet} backbone configured with four stages of channel dimensions $[64,64,64,16]$. Number of blocks at each stage is set to $[4,2,2,2]$ or $[2,2,2,2]$ to match the resolutions of input TSDF-Def tensors and output embedded features. A lightweight 
$1\times1\times1$ convolutional projection follows, mapping the encoder output channels to the latent embedding dimension.

The decoder mirrors the encoder’s hierarchical structure but operates in reverse. Following NeCGS~\cite{ren2024necgs}, we build the auto-decoder with a head layer, which increases the channel of its input, and L cascaded upsampling modules,  which progressively upscale the feature tensor. 3D PixelShuffle~\cite{shi2016real} layers are added between the convolution and activation layers to achieve upscaling. 
Three decoder variants are designed to support different output resolutions: 64, 128, and 256. All three share the same encoder and differ only in their decoder channel widths. For the 64-resolution model, the decoder channel configuration is $[32,24,16,12,8]$, for the 128-resolution model, it is
$[48,36,24,16,12]$, and for the 256-resolution model, it is $[64,48,32,24,16]$. We scale this with embedded feature dimensions progressively to ensure the model is lightweight while powerful enough for 4D mesh compression. The decoder terminates with a final $3\times3\times3$ quantized convolution layer producing four output channels corresponding to the voxel reconstruction.

We use an initial learning rate of 0.001, a cosine annealing learning rate scheduler with a minimum of 1e-5, a warmup proportion of 0.2 for linear warmup, and 8-bit quantization throughout quantized layers. The activation function used throughout is GELU~\cite{hendrycks2016gaussian}. All the experiments are conducted using a Lambda Vector equipped with an AMD Ryzen Threadripper 7960X 24-core 4.20 GHz CPU and an RTX 4090 GPU with 24 GB of memory.

\paragraph{Latent Mapping Module} For simplicity, we replace MLP layers in PointNet~\cite{qi2017pointnet} with 3 convolution layer with dimension 128 and kernel size 1, to make the output feature dimension compact.

The sinusoidal time encoder uses 16 frequency bands. Each frequency contributes both sine and cosine terms, resulting in a 32-dimensional time embedding.

The fusion MLP in the LatentMapperPointNet consists of three fully connected layers with GELU activations, mapping a 544-dimensional input to a compact 32-dimensional latent vector. The input dimension (544) comes from concatenating four 128-dimensional PointNet features (start, end, intermediate, and their difference) and a 32-dimensional sinusoidal time embedding derived from 16 frequency bands. The MLP layers are structured as $(544,256,256,32)$, where the first two layers use GELU~\cite{hendrycks2016gaussian} activations, and the final layer outputs the latent code. A LayerNorm~\cite{ba2016layer} operation is applied to the 32-dimensional output, followed by a learnable scaling factor (initialized to 0.1) that stabilizes training by controlling output magnitude.

\paragraph{3D interpolation Module}
The 3D interpolation Module begins with an input voxel encoder, which is a two-layer MLP with GELU~\cite{hendrycks2016gaussian} activations that maps per-voxel input features with 16 channels to a higher-dimensional latent space of 64 features. These voxel features are projected to a smaller transformer space of $x$ dimensions through another two-layer QuantLinear MLP. We set $x$ to 16, 24, and 32 for the resolutions of 64, 128, and 256, respectively. Both encoder and decoder transformers operate on this $x$ dimensional latent space with 4 attention heads, a feedforward expansion ratio of $4\times$, and 2 encoder plus 2 decoder layers each. Spatial information is injected through coordinate embeddings of the $8\times8\times8$ voxel grid, projected via a linear layer to match the transformer dimension. Temporal context is encoded by a sinusoidal time embedding with 16 frequency bands. For conditioning, each intermediate step uses a latent vector of dimension 32, fused with the time embedding and processed by a two-layer MLP.



\cref{alg:compression} summarizes the whole compression and decompression process.
\begin{algorithm}[]
\caption{N4MC Compression and Decompression}
\label{alg:compression}
\begin{algorithmic}[1]
\Procedure{Compress}{$\{\mathbf{M}_i\}_{i=1}^N$}
\State Generate volume centers, denoted as $\mathcal{C}=\{\mathbf{C}_i\}_{i=1}^{N}$
    \For{$i = 1$ \textbf{to} $N$}
        \State $\mathbf{T}_i \gets \text{TSDF-Def}(\mathbf{M}_i)$ \Comment{Generate TSDF-Def tensor}
        \State $\mathbf{F}_i \gets E_\theta(\mathbf{T}_i)$ \Comment{Encode to embedded features}
        \State $\mathcal{Q}(\mathbf{F}_i)$ \Comment{Quantize features}
    \EndFor
    \State $\mathbf{Z}=f_\theta(\mathcal{C})$ \Comment{Generate latent codes}
    \State Entropy coding
\EndProcedure
\vspace{0.5em}
\Procedure{Decompress}{$\text{bitstreams}$}
    \State Decode bitstreams to recover $\mathcal{F}= \{\mathbf{F_i}\}^{N}_{i=1}$ and $\mathbf{Z}$
    \For{$t = 1$ \textbf{to} $N$} 
        \State $\hat{\mathbf{F}}_t \gets \mathcal{I}_{\mathcal{Q}(\phi)}(\mathcal{F}, \mathbf{Z}, t)$ \Comment{Interpolate embedded features}
        \State $\hat{\mathbf{T}}_t \gets D_{\mathcal{Q}(\psi)}(\hat{\mathbf{F}}_t)$ \Comment{Decode TSDF-Def tensor}
        \State $\hat{\mathbf{M}}_t \gets \text{DMC}(\hat{\mathbf{T}}_t)$ \Comment{Reconstruct mesh~\cite{ren2024necgs}}
    \EndFor
    \State \textbf{return} $\{\hat{\mathbf{M}}_t\}_{t=1}^N$
\EndProcedure
\end{algorithmic}
\end{algorithm}

%

\section{Additional Results}
In this section, we first provide additional quantitative and qualitative results on our main benchmark, which includes four MPEG mesh sequences: "Dancer", "Basketball player", "Mitch", and "Thomas". We then present qualitative results on the "Mixed" dataset. Next, we evaluate N4MC on our own custom captured sequences featuring everyday human activities and object interactions, as well as on high-variability synthetic data from Thingi10K~\cite{Thingi10K}. Finally, we include failure cases with volume center latent mapping disabled as a supplemental ablation study.

Playback results of the decoded mesh sequences for N4MC and all baselines are provided in the supplementary video.
\label{sec:addresults}

\begin{figure*}[t]
    \centering
    \includegraphics[width=1\linewidth]{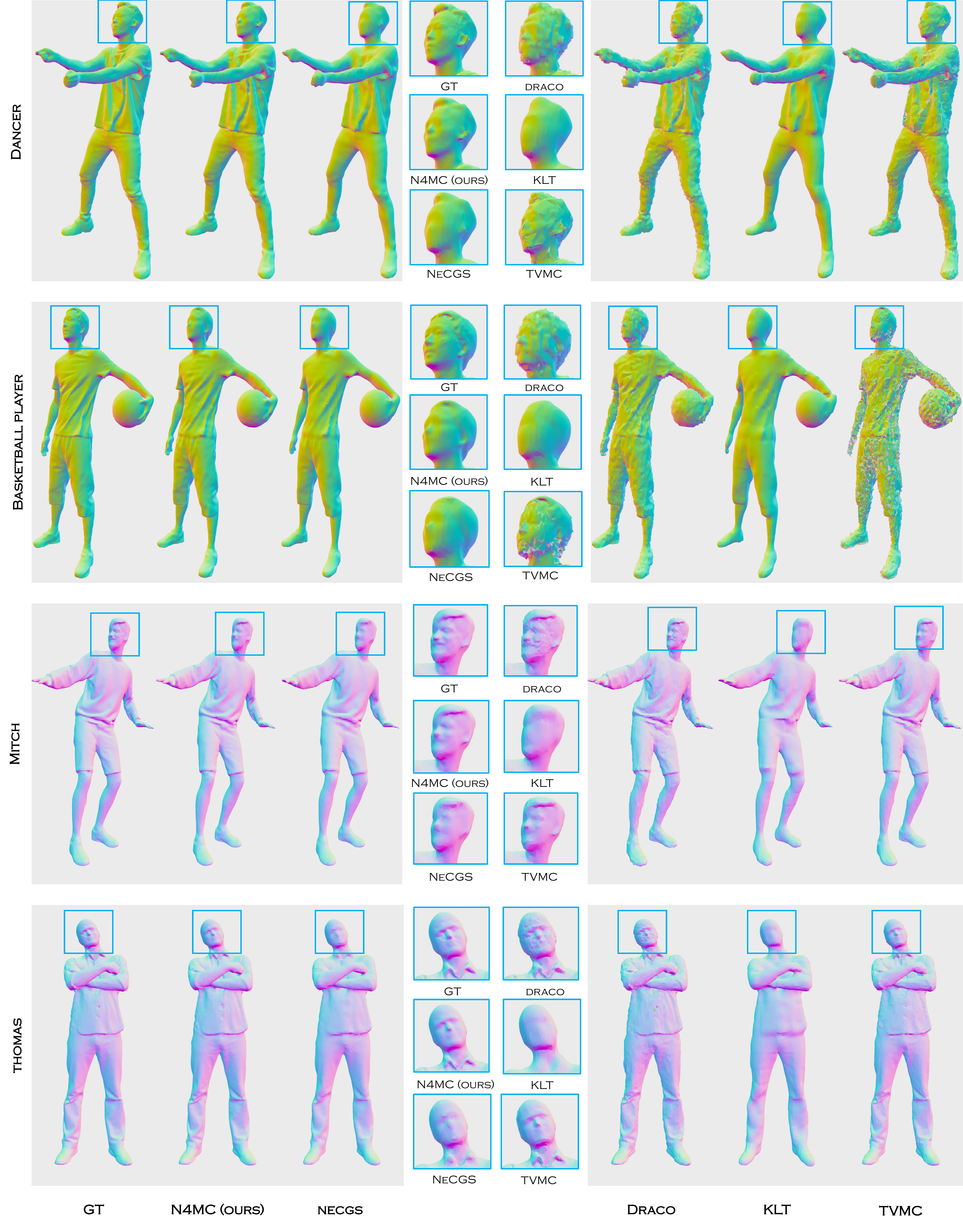} 
    \caption{Complete qualitative comparison of our N4MC with baselines at a bitrate around 4 Mbps.} 
    \label{fig:visual_comparison_full}
\end{figure*}

\paragraph{Complete Comparison} We present the full qualitative comparison results in \cref{fig:visual_comparison_full}, corresponding to the simplified version shown in \cref{fig:visual_comparison}. The complete meshes further demonstrate that N4MC outperforms all baselines in visual quality at a similar bitrate of around 4 Mbps. The improvements are evident not only in the zoomed-in head region but also in other challenging areas, such as the hands in the "Basketball player" sequence.

\begin{figure*}[]
     \centering
     \begin{subfigure}[b]{0.24\textwidth}
         \centering
         \includegraphics[width=1\textwidth]{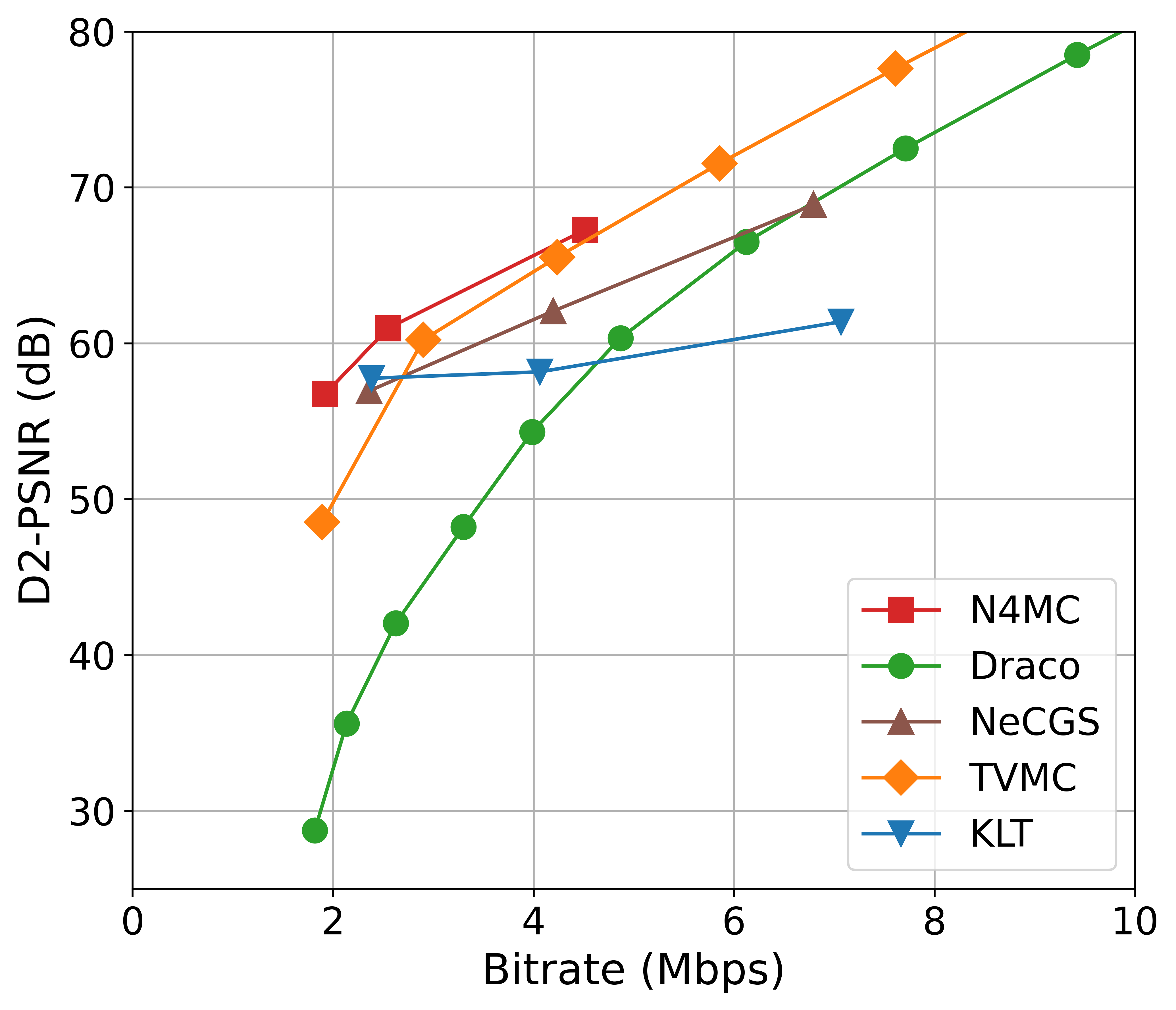}
         \caption{Comparison of D2-PSNR vs. bitrate on "Dancer"}
         \label{fig:compare:d2-psnr-a}
     \end{subfigure}
     \hfill
     \begin{subfigure}[b]{0.24\textwidth}
         \centering
         \includegraphics[width=1\textwidth]{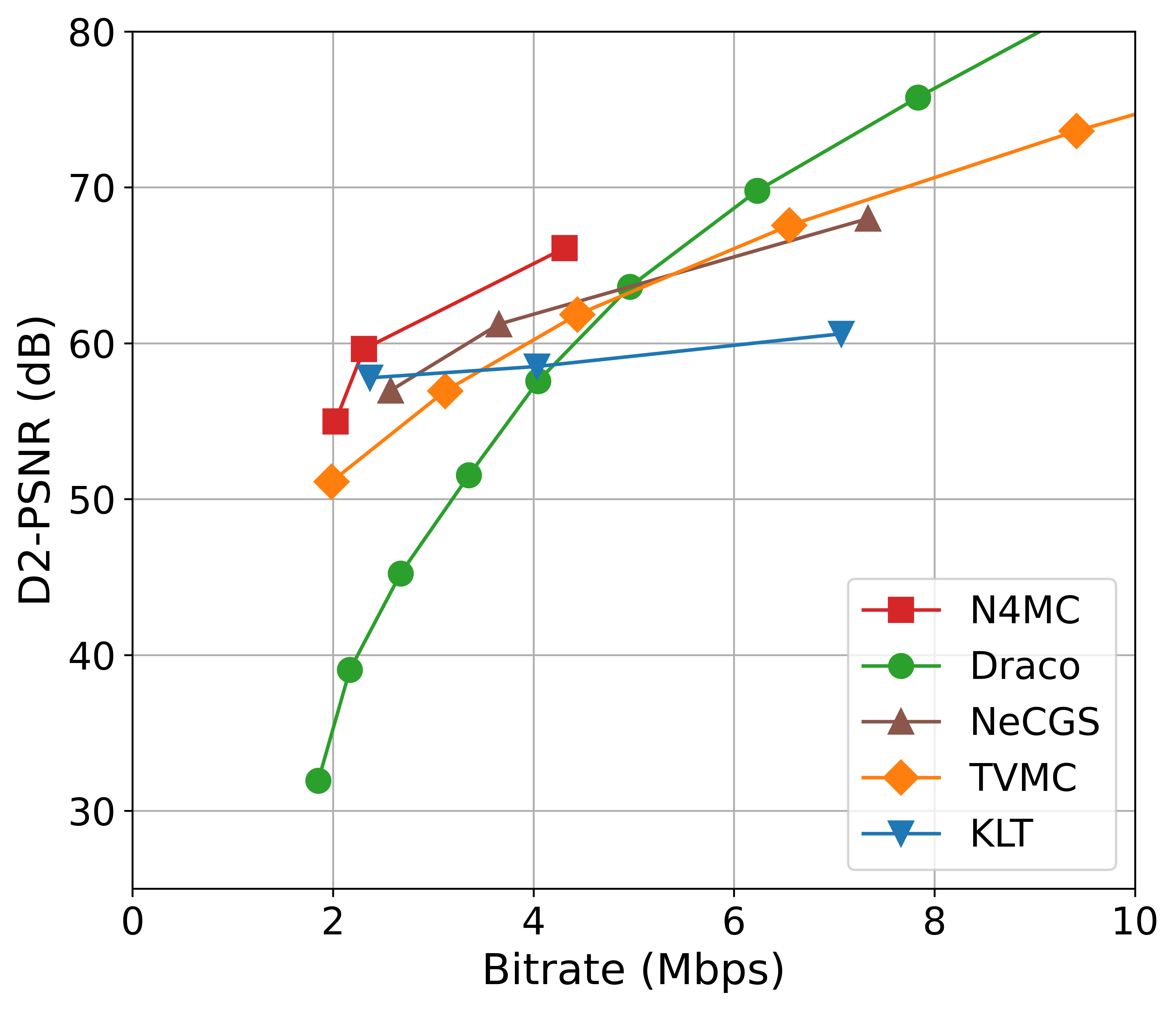}
         \caption{Comparison of D2-PSNR vs. bitrate on "Basketball player"}
         \label{fig:compare:d2-psnr-b}
     \end{subfigure}
     \hfill
     \begin{subfigure}[b]{0.24\textwidth}
         \centering
         \includegraphics[width=1\textwidth]{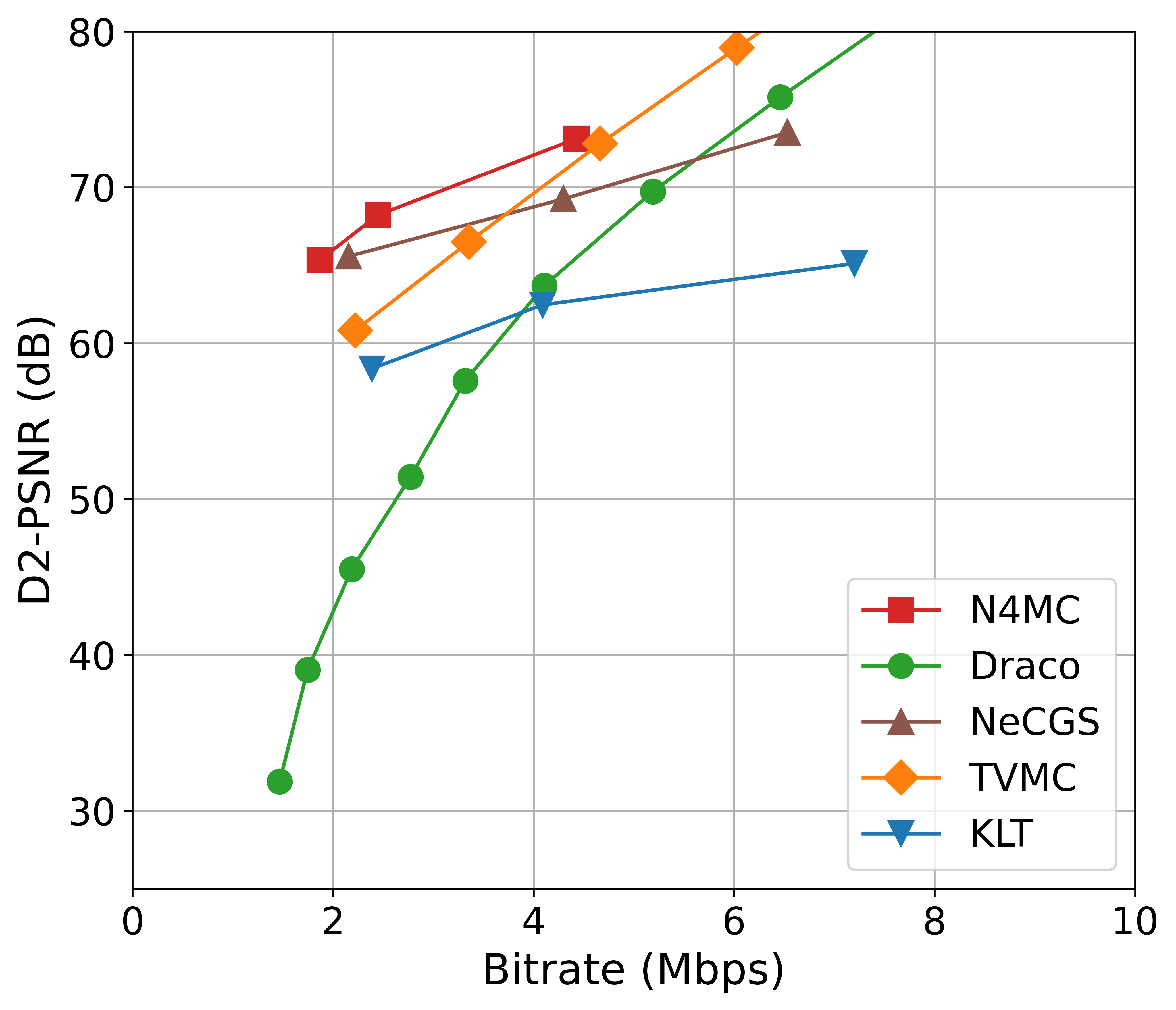}
         \caption{Comparison of D2-PSNR vs. bitrate on "Mitch"}
         \label{fig:compare:d2-psnr-c}
     \end{subfigure}
     \hfill
     \begin{subfigure}[b]{0.24\textwidth}
         \centering
         \includegraphics[width=1\textwidth]{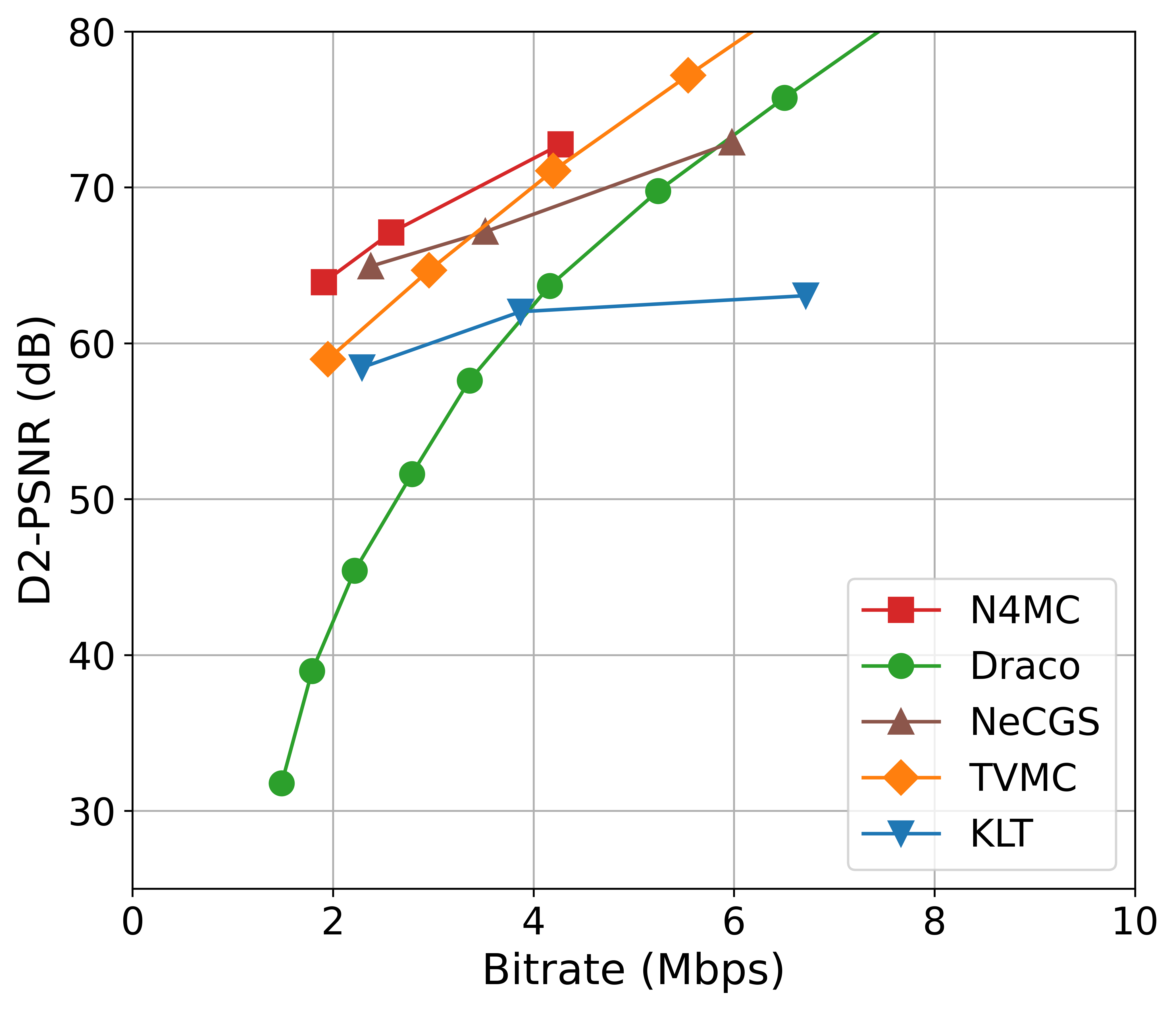}
         \caption{Comparison of D2-PSNR vs. bitrate on "Thomas"}
         \label{fig:compare:d2-psnr-d}
     \end{subfigure}
        \caption{Objective rate-distortion (RD) performance comparison of geometry-based \textbf{D2-PSNR} versus bitrate on four 4D mesh sequences: "Dancer", "Basketball player", "Mitch", and "Thomas". To get the target bitrates, we use the same setting as in~\cref{RD-performance}.}
    \label{RD-performance-d2-PSNR}
\end{figure*}

\begin{figure*}[]
     \centering
     \begin{subfigure}[b]{0.24\textwidth}
         \centering
         \includegraphics[width=1\textwidth]{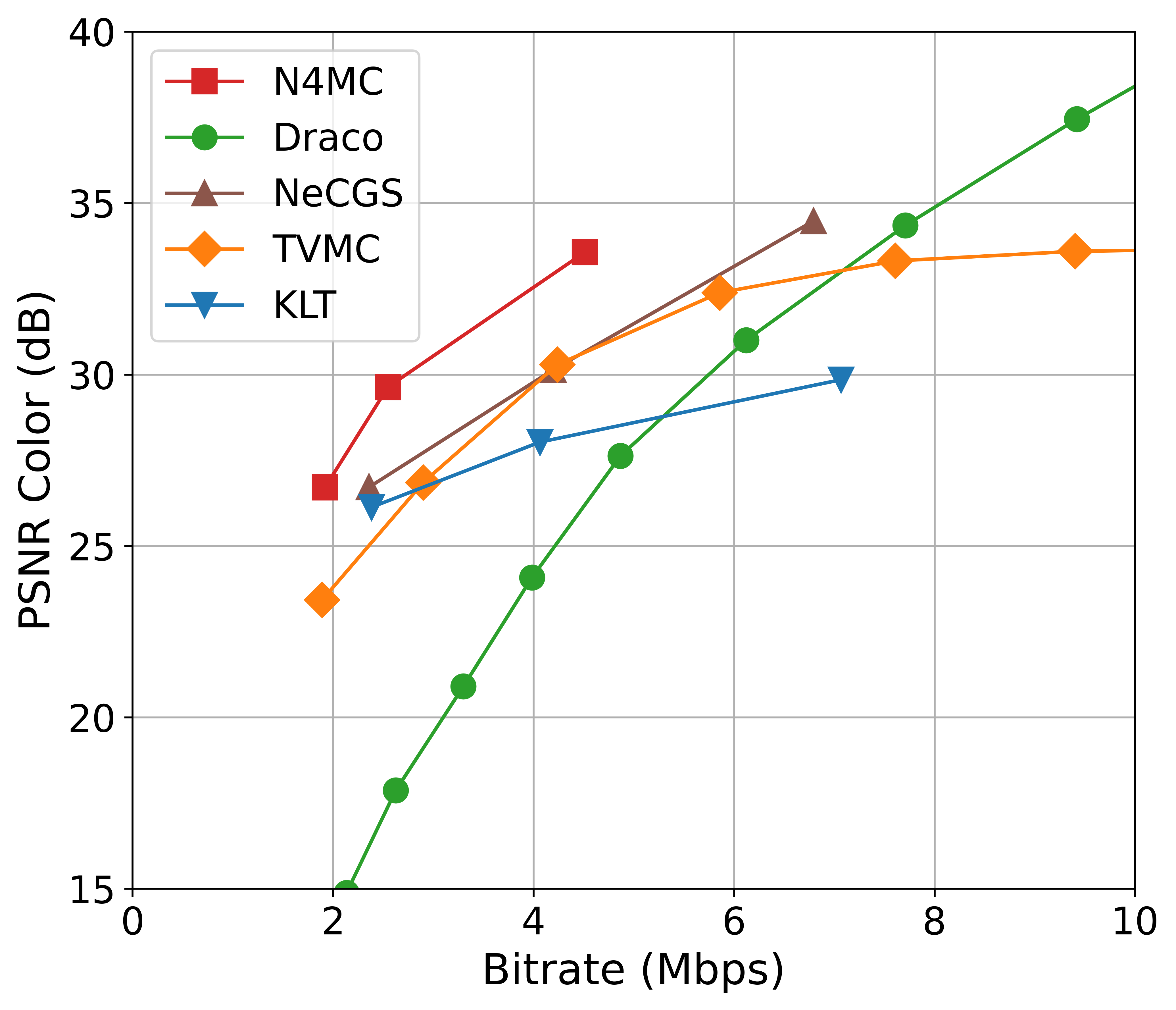}
         \caption{Comparison of PSNR vs. bitrate on "Dancer".}
         \label{fig:compare:psnr-a}
     \end{subfigure}
     \hfill
     \begin{subfigure}[b]{0.24\textwidth}
         \centering
         \includegraphics[width=1\textwidth]{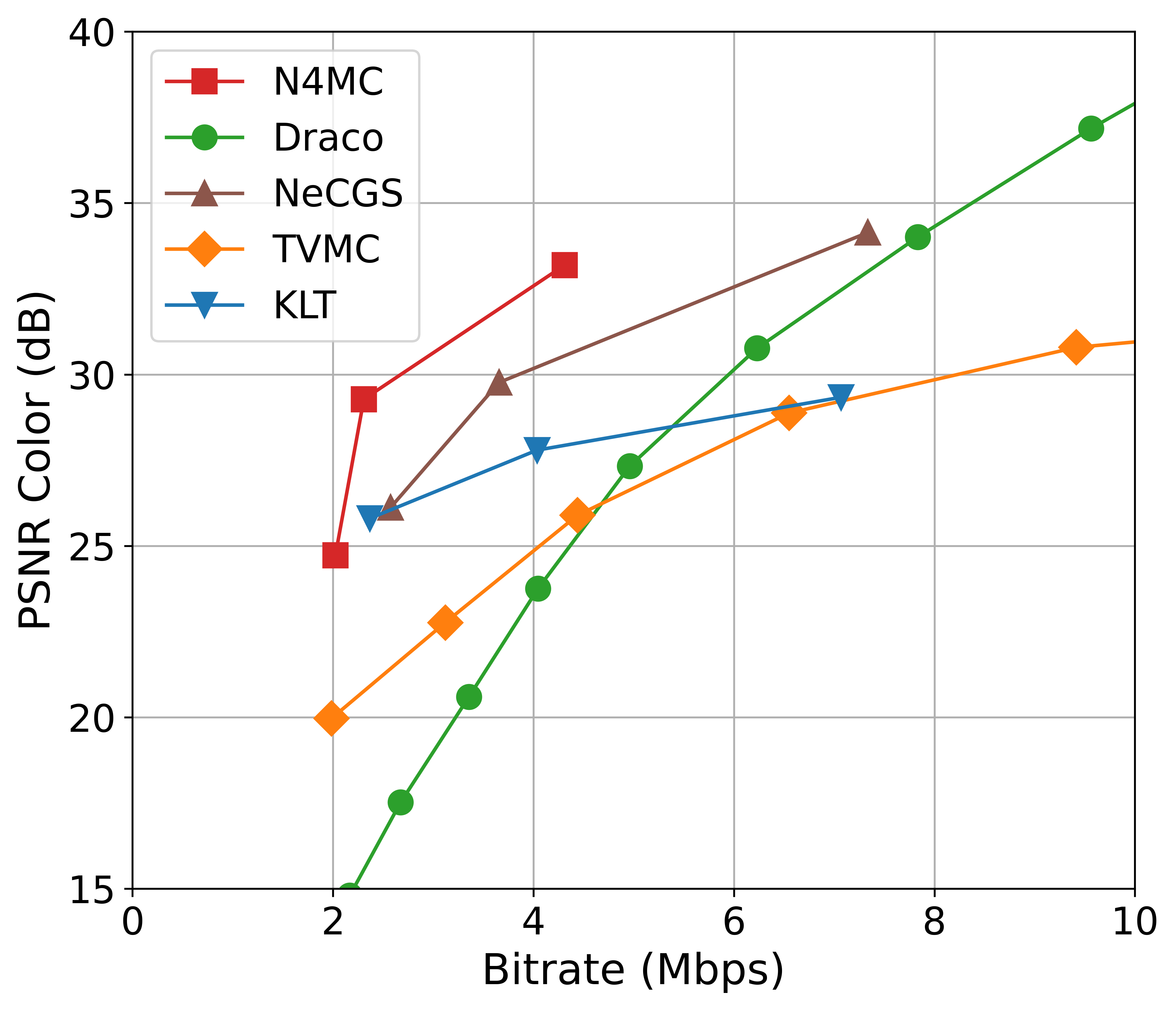}
         \caption{Comparison of PSNR vs. bitrate on "Basketball player"}
         \label{fig:compare:psnr-b}
     \end{subfigure}
     \hfill
     \begin{subfigure}[b]{0.24\textwidth}
         \centering
         \includegraphics[width=1\textwidth]{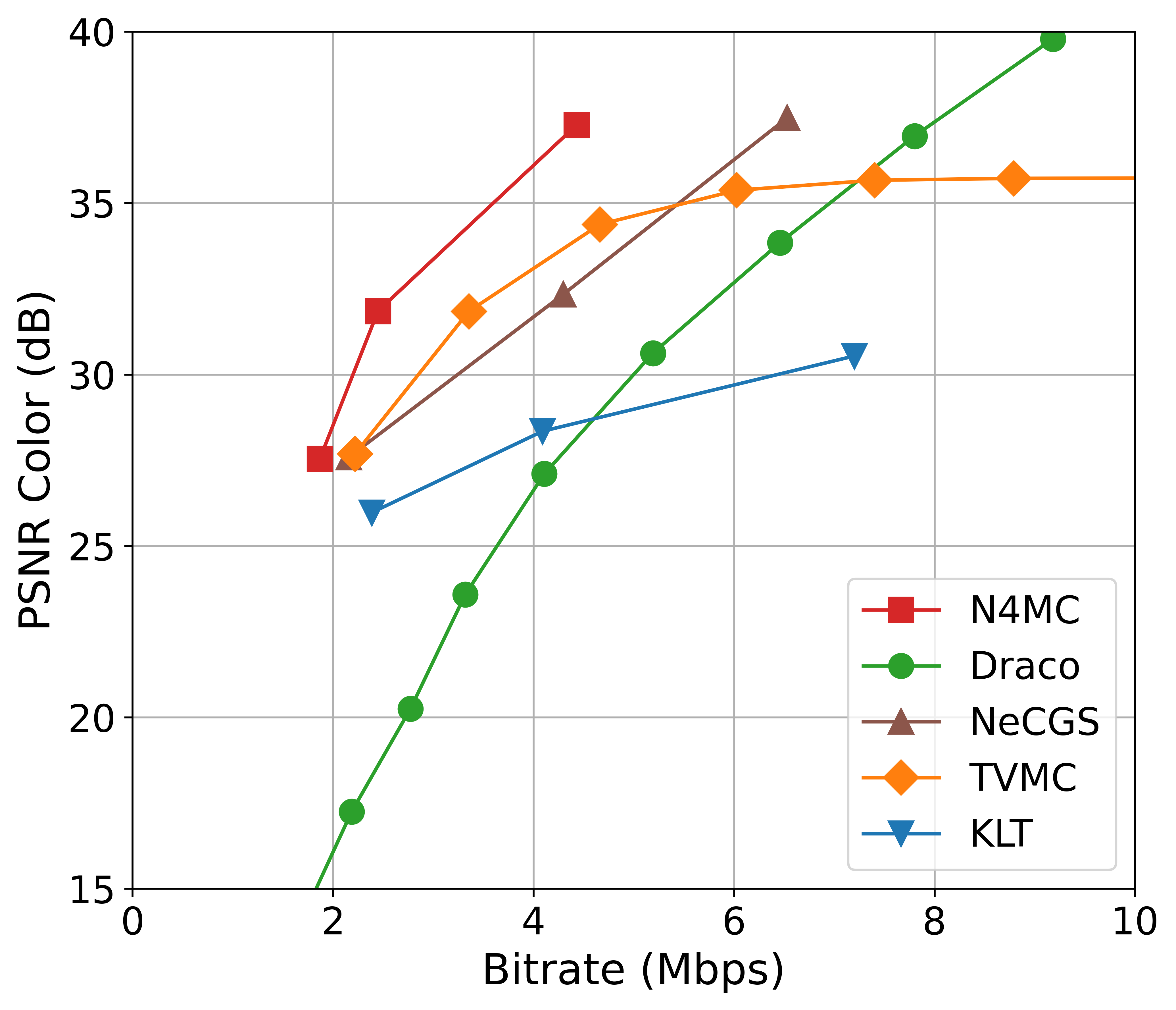}
         \caption{Comparison of PSNR vs. bitrate on "Mitch"}
         \label{fig:compare:psnr-c}
     \end{subfigure}
     \hfill
     \begin{subfigure}[b]{0.24\textwidth}
         \centering
         \includegraphics[width=1\textwidth]{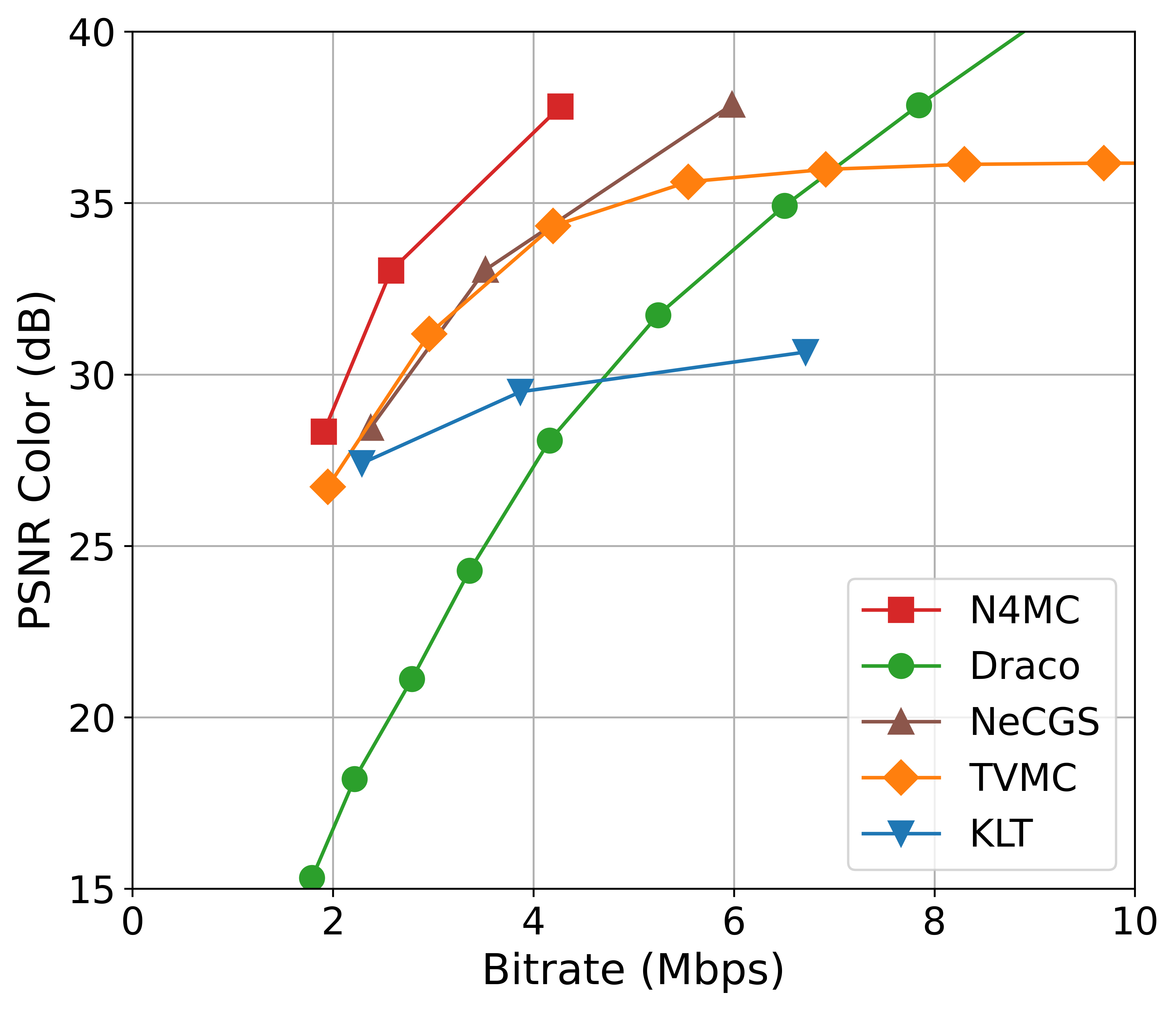}
         \caption{Comparison of PSNR vs. bitrate on "Thomas"}
         \label{fig:compare:psnr-d}
     \end{subfigure}
        \caption{Objective rate-distortion (RD) performance comparison of image-based \textbf{PSNR} versus bitrate on four 4D mesh sequences: "Dancer", "Basketball player", "Mitch", and "Thomas". To get the target bitrates, we use the same setting as in~\cref{RD-performance}.}
    \label{RD-performance-PSNR}
\end{figure*}

Additional rate-distortion performance using geometry-based D2-PSNR and image-based PSNR is provided in~\cref{RD-performance-d2-PSNR} and~\cref{RD-performance-PSNR}, respectively. N4MC consistently achieves the best performance across both metrics, highlighting its superior capability in 4D mesh compression.

\begin{figure}[]
    \centering
    \includegraphics[width=\linewidth]{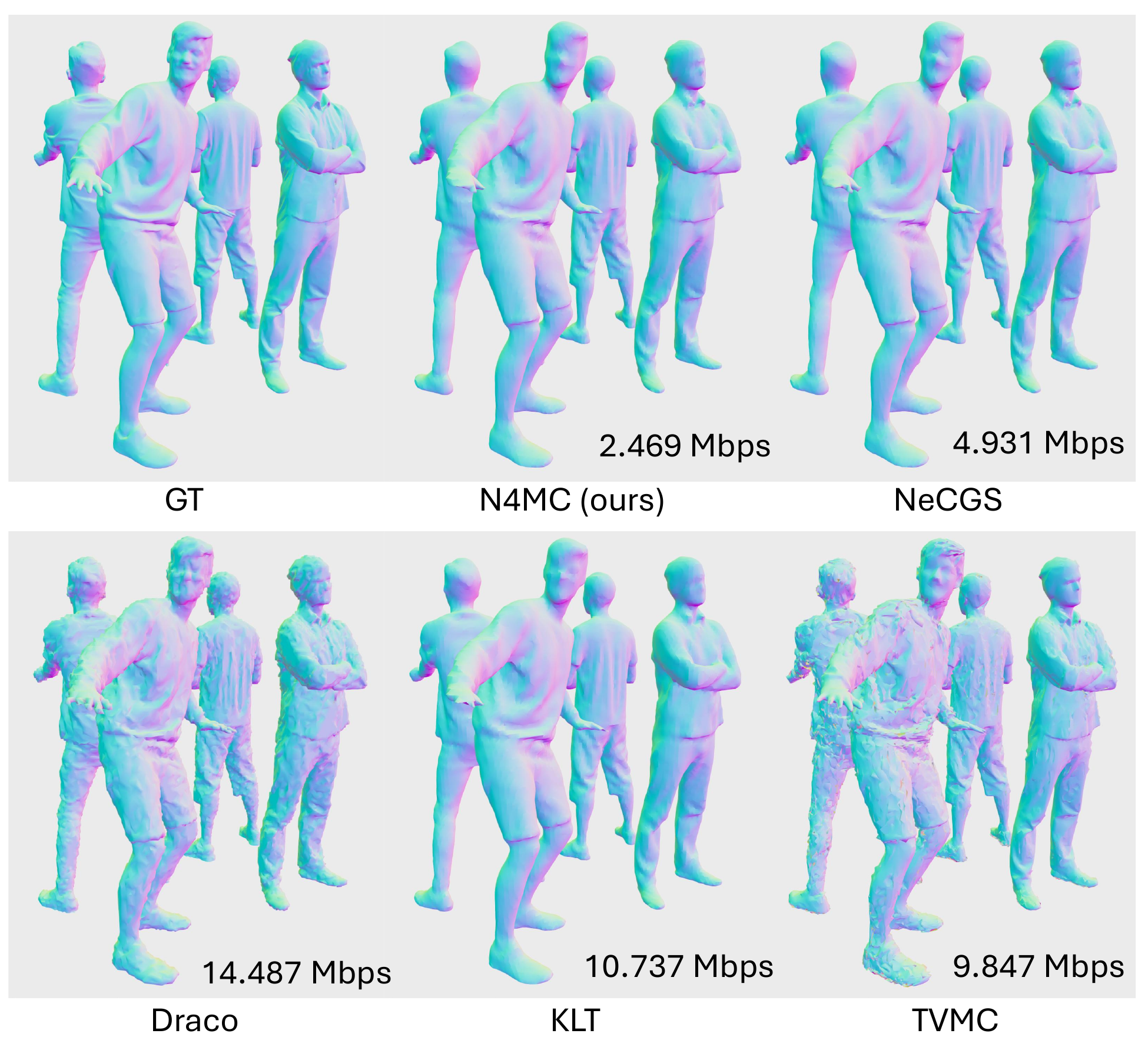} 
    \caption{Qualitative comparison of our N4MC with baselines on "Mixed" dataset.} 
    \label{fig:visual_comparison-mixed}
\end{figure}

\paragraph{Qualitative Results for "Mixed" dataset}
To demonstrate the scalability of N4MC, we evaluate it on the "Mixed" dataset, which consists of 4 MPEG mesh sequences. The qualitative comparisons of N4MC and all baselines corresponding to~\cref{tab:comparison_mixed} are shown in~\cref{fig:visual_comparison-mixed}. For fair comparison, we set the resolution of NeCGS~\cite{ren2024necgs} and N4MC to 128, use a quantization parameter of $qp=7$ for Draco~\cite{Draco2024}, $qp=5$ for TVMC~\cite{chen2025tvmc}, and 128 basis vectors for KLT~\cite{realtime2018}. These settings produce comparable reconstruction quality around 60 dB D2-PSNR. 

With the lowest bitrate of 2.469 Mbps, N4MC achieves nearly the same visual quality as NeCGS~\cite{ren2024necgs} at 4.931 Mbps, while maintaining facial expressions and avoiding noticeable distortions. In contrast, KLT~\cite{realtime2018} requires over 10 Mbps to achieve a similar quality. TVMC~\cite{chen2025tvmc} and Draco~\cite{Draco2024} still suffer from visible artifacts on faces and thin structures such as arms and hands, despite requiring more than $4\times$ the bitrate.

\paragraph{Experiments on Real Captured Data} 

To evaluate N4MC on a broader range of time-varying scene mesh sequences, we built a high-resolution capture setup with four Azure Kinect cameras~\cite{AzureKinectDK}, recording depth frames at $640\times576$ resolution and 30 FPS. Using this setup, we use Open3D~\cite{Zhou2018} to convert the RGB-D images into TSDF volumes and then extract meshes via the Marching Cubes algorithm~\cite{lorensen1998marching}.

\begin{figure}[]
    \centering
    \includegraphics[width=0.9\linewidth]{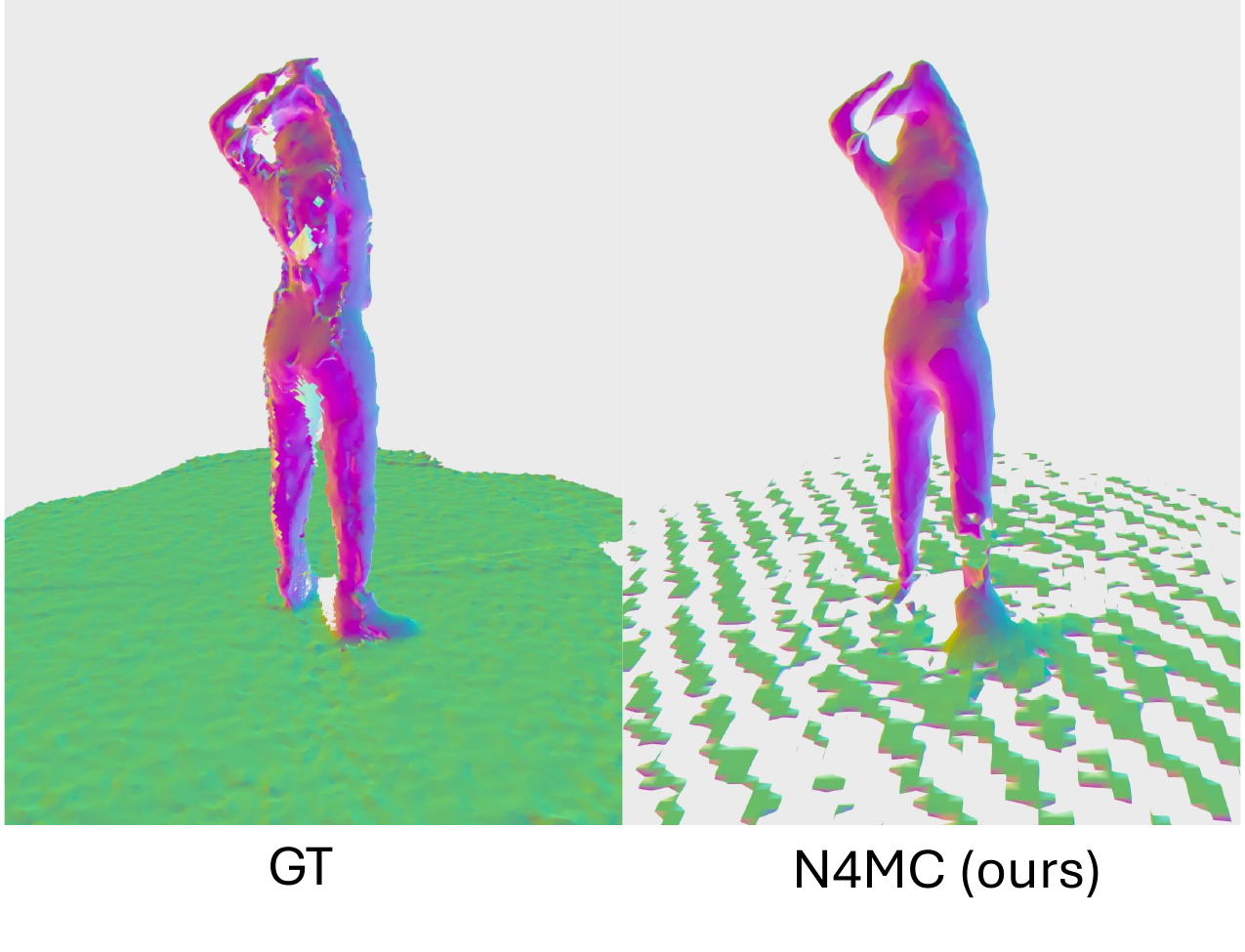} 
    \caption{Qualitative comparison of our N4MC with baselines on a real captured dataset.} 
    \label{fig:visual_comparison-arena4}
\end{figure}

In~\cref{fig:visual_comparison-arena4}, we present qualitative results of ground truth and N4MC decoded mesh on one of our real captured 4D mesh sequences called "Arena", with the setting of N4MC resolution to 128. Due to camera noise and the limited four-camera setup, the meshes directly extracted from RGB-D images (GT) exhibit noticeable distortions. Nevertheless, N4MC operates on these raw sequences without any pre-processing and still produces high-quality reconstructions, demonstrating its practicality and robustness. 

One remaining artifact is the noisy and jittery appearance of the floor in the N4MC results. This occurs because the floor is essentially a flat plane, and the Marching Cubes algorithm attempts to generate a closed and watertight surface during mesh extraction.

\paragraph{Experiments on Synthetic Data} We evaluate N4MC on synthetic data from Thingi10K~\cite{Thingi10K}. We filter meshes with vertex counts between 20,000 and 30,000 and randomly sample 100 of them to form a sequence for evaluation. Since Thingi10K contains static meshes rather than 4D mesh sequences, we disable the 3D interpolation module in N4MC during this experiment.

\begin{figure*}[]
    \centering
    \includegraphics[width=0.9\linewidth]{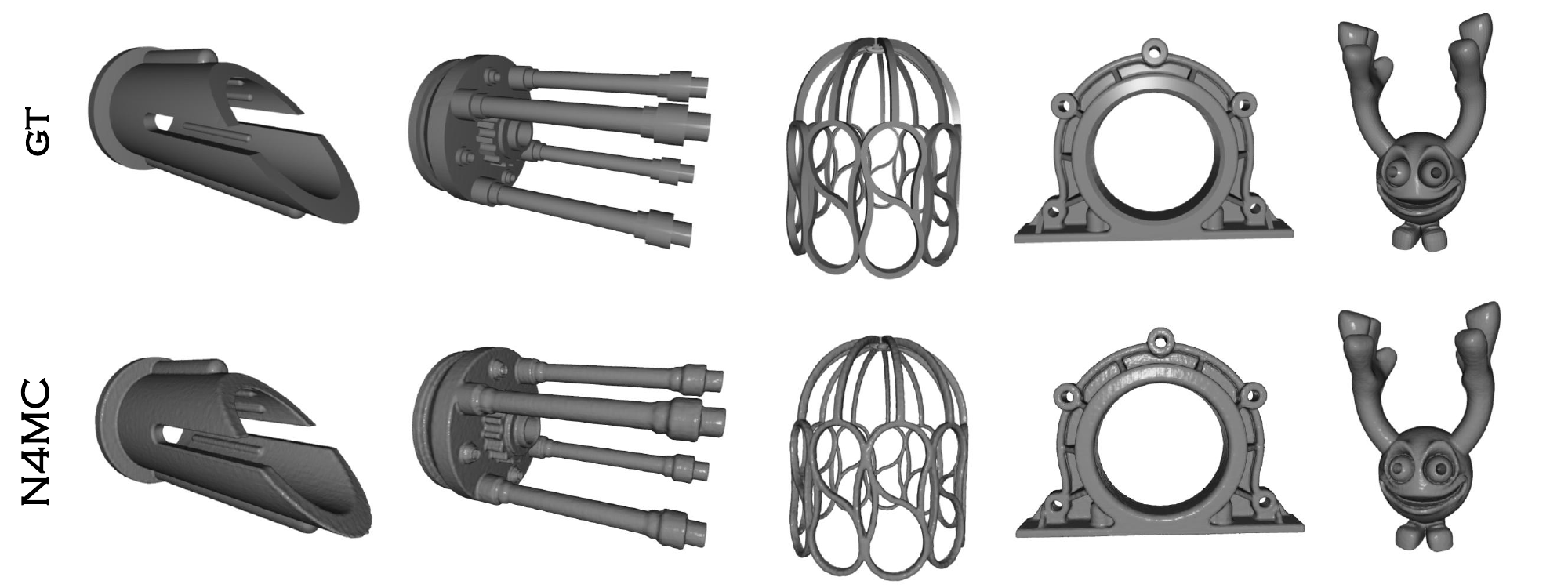} 
    \caption{Qualitative results of our N4MC on synthetic dataset Thingi10K~\cite{Thingi10K}.} 
    \label{fig:thingi10k}
\end{figure*}

~\cref{fig:thingi10k} presents qualitative reconstruction results obtained by N4MC at a resolution of 128, achieving a compression ratio (CR) of 89.56. We also compare N4MC against Neural Progressive Meshes~\cite{chen2023neural} on Thingi10K using the metrics $d_\text{pm}$ (mean point-to-mesh distance) and $d_\text{normal}$ (average normal error), following the definitions in~\cite{chen2023neural}. Since no official implementation of Neural Progressive Meshes is publicly available, we reference the results reported in their paper.

With the interpolation module disabled, N4MC achieves a CR of 89.56, approximately $1.45\times$ the performance of Neural Progressive Meshes, while maintaining comparable reconstruction quality. The mean point-to-mesh distance and average normal error increase slightly from 4.12 to 10.42, and from 7.19$\degree$ to 9.26$\degree$, respectively.

\begin{table}[h]
\centering
\caption{Comparisons of N4MC and Neural Progressive Meshes~\cite{chen2023neural} on Thingi10K. $d_\text{pm}$ represents the mean point-to-mesh distance. $d_\text{normal}$ represents the average normal error.}
\resizebox{1\linewidth}{!}{
\begin{tabular}{lccc}
\toprule
Method & $d_\text{pm} (\times10^{-4}) \downarrow$ & $d_\text{normal} \downarrow$ & CR $\uparrow$\\
\midrule
NPM~\cite{chen2023neural} & 4.12 & 7.19$\degree$ & 61.39 \\
N4MC (ours)   & 10.42 & 9.26$\degree$ & 89.56 \\
\bottomrule
\end{tabular}
}
\label{tab:thingi10k}
\end{table}

\begin{figure}[]
    \centering
    \includegraphics[width=\linewidth]{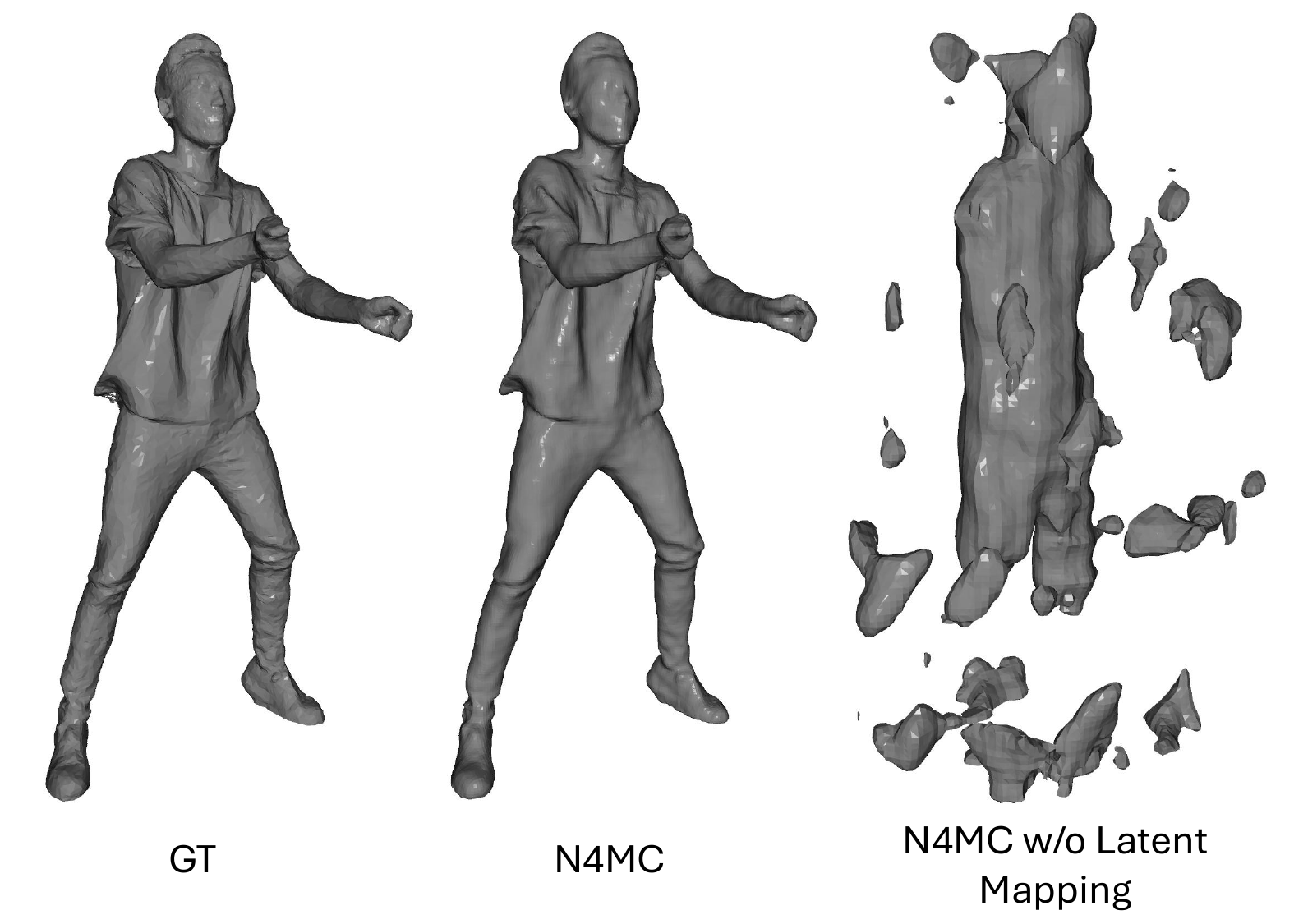} 
    \caption{Ablation on volume center latent mapping. Without latent mapping, N4MC could fail.} 
    \label{fig:ablation_latent}
\end{figure}

\paragraph{Failure Cases Without Volume Center Latent Mapping}
In 2D image interpolation–based video compression methods, obtaining appropriate latent codes is crucial for resolving interpolation ambiguities. To analyze this effect in N4MC, we perform an additional ablation study by disabling the proposed volume-center latent mapping module, as shown in~\cref{fig:ablation_latent}. Without this module, the transformer fails to learn meaningful intermediate representations, resulting in poor interpolation and incorrect mesh reconstruction.

\section{Additional Results of N4MC Decoder on VR headset}
\begin{figure*}[]
    \centering
    \includegraphics[width=0.9\linewidth]{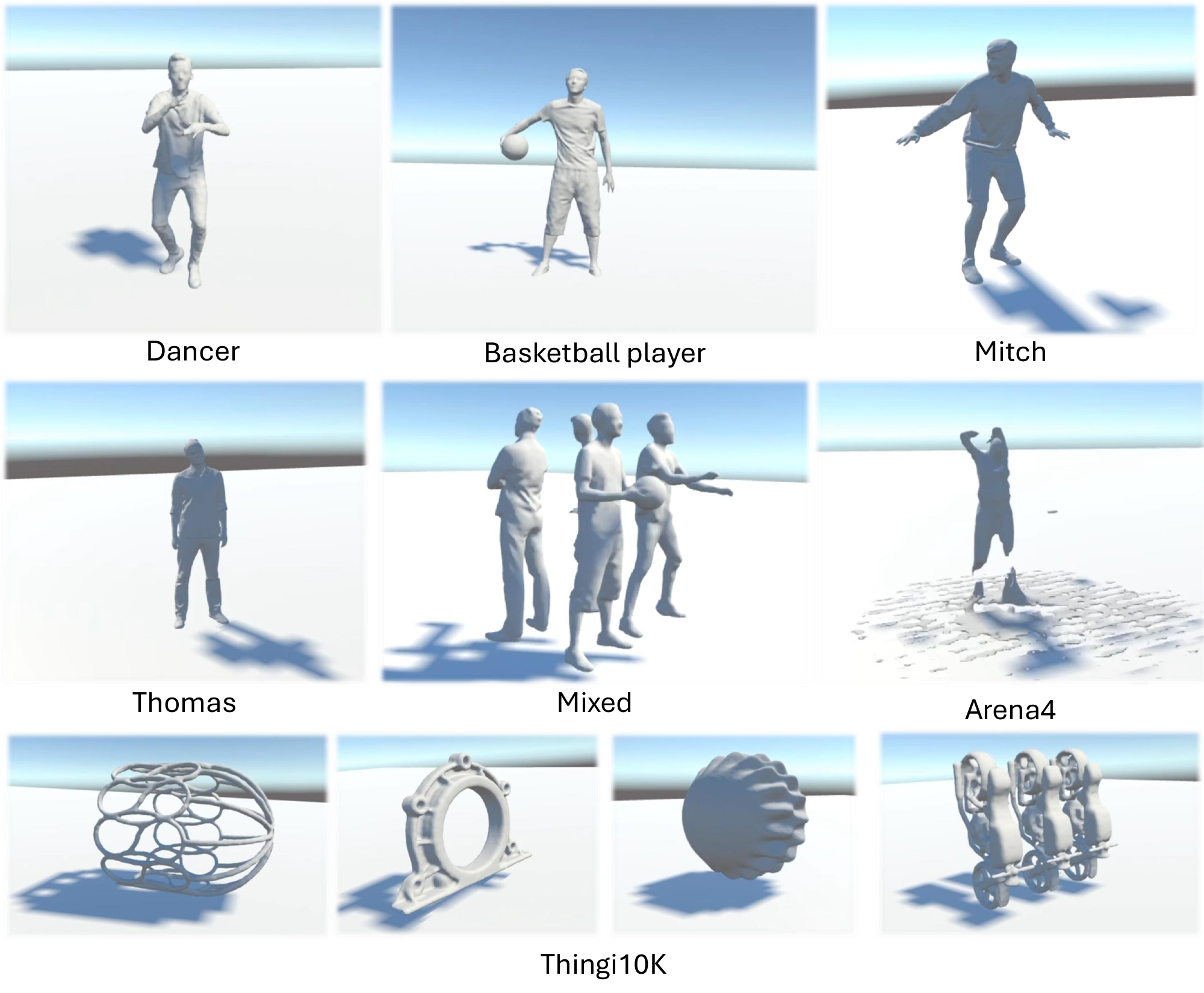} 
    \caption{Decoded meshes in Meta Quest 3.} 
    \label{fig:vr_results}
\end{figure*}

In~\cref{experiments}, we've provided N4MC's decoding time on multiple mobile devices, including Meta Quest 3, Pixel 9, and Xiaomi 12S Pro. Here we show additional visual results of N4MC decoded meshes and the playback system in Meta Quest 3 in~\cref{fig:vr_results}.

\section{Limitations and Future Work}
The current N4MC exhibits overfitting, as it compresses 4D meshes into two bitstreams: compressed embedding features and compressed model parameters, which results in limited generalization to unseen data. In future work, N4MC could be extended to train on larger-scale datasets and serve as a foundation model that can be deployed across diverse devices. To further reduce GPU memory usage and accelerate training, the uniform TSDF representation could be replaced with a pruned version that retains only near-surface voxels.

\end{document}